\documentclass[10pt,twocolumn,letterpaper]{article}

\usepackage[accsupp]{axessibility}
\usepackage[pagenumbers]{iccv} 

\definecolor{iccvblue}{rgb}{0.21,0.49,0.74}
\usepackage[pagebackref,breaklinks,colorlinks,allcolors=iccvblue]{hyperref}

\def\paperID{6530} 
\def\confName{ICCV}
\def\confYear{2025}

\usepackage{array}
\newcolumntype{P}[1]{>{\centering\arraybackslash}p{#1}}
\newcolumntype{L}[1]{>{\left\arraybackslash}p{#1}}
\usepackage{subcaption}
\usepackage{amsmath}
\usepackage{booktabs}
\usepackage{multirow}
\usepackage{amssymb}
\usepackage{makecell}

\usepackage[dvipsnames]{xcolor}
\usepackage{hhline}
\usepackage{colortbl}

\usepackage{color}
\definecolor{ao(english)}{rgb}{0.0, 0.5, 0.0}
\newcommand{\BLUE}[1]{{\color{blue}#1}}
\newcommand{\RED}[1]{{\color{red}#1}}
\newcommand{\GREEN}[1]{{\color{ao(english)}#1}}

\newcommand{\SUPP}[0]{\textit{Supplementary}}
 \newcommand*{\x}
 {\!\times\!}
\newcommand{\B}[1]{{\textbf{\underline{#1}}}}
\newcommand{\SB}[1]{{\underline{#1}}}
\definecolor{mygray}{gray}{.9}

\newcommand{\IMPROV}[1]{{\color{ao(english)}\textbf{\textsuperscript{#1}}}
}
\newcommand{\comm}[1]
{ \iffalse \fi }
\newcommand{\R}{\mathbb{R}}
\definecolor{mypink}{RGB}{255, 20, 147}
\hypersetup{
    colorlinks=true,
    linkcolor=blue,
    urlcolor=mypink,
}

\title{ Colors See Colors Ignore: Clothes Changing ReID with Color Disentanglement
\vspace{-10pt}
}

\author{
Priyank Pathak \qquad Yogesh S. Rawat \vspace{.5em} \\
Center for Research in Computer Vision, University of Central Florida \\
{\tt\normalsize \{priyank, yogesh\}@ucf.edu}\\
\normalsize\textbf{\url{https://ucf-crcv.github.io/ReID/CSCI}} 
\vspace{-2pt}
}

\begin{document}
\maketitle
\begin{abstract}
Clothes-Changing Re-Identification (CC-ReID) aims to recognize individuals across different locations and times, irrespective of clothing. 
Existing methods often rely on additional models or annotations to learn robust, clothing-invariant features, making them resource-intensive. 
In contrast, we explore the use of color—specifically foreground and background colors—as a lightweight, annotation-free proxy for mitigating appearance bias in ReID models.
We propose \textbf{Colors See, Colors Ignore (CSCI)}, an RGB-only method that leverages color information directly from raw images or video frames. CSCI efficiently captures color-related appearance bias (\textbf{`Color See'}) while disentangling it from identity-relevant ReID features (`\textbf{Color Ignore}'). To achieve this, we introduce \textbf{S2A self-attention}, a novel self-attention to prevent information leak between color and identity cues within the feature space. Our analysis shows a strong correspondence between learned color embeddings and clothing attributes, validating color as an effective proxy when explicit clothing labels are unavailable.
We demonstrate the effectiveness of CSCI on both image and video ReID with extensive experiments on four CC-ReID datasets. 
We improve baseline by Top-1 2.9\% on LTCC and 5.0\% on PRCC for image-based ReID, and 1.0\% on CCVID and 2.5\% on MeVID for video-based ReID without relying on additional supervision. 
Our results highlight the potential of color as a cost-effective solution for addressing appearance bias in CC-ReID.
\textit{Github:} \textbf{\url{https://github.com/ppriyank/ICCV-CSCI-Person-ReID}}.    
\vspace{-5pt}
\end{abstract}    
\section{Introduction}
\label{sec:intro}

Person Re-Identification (ReID) aims to recognize individuals across different cameras, times, and locations based on visual cues. While traditional ReID~\cite{pathak2020fine, pathak2020video} assumes consistent clothing, real-world scenarios often involve changes in attire, making Clothes-Changing ReID (CC-ReID)~\cite{gu2022clothes} a more challenging yet practical task. In CC-ReID, the model must distinguish identities despite drastic variations in appearance, while also handling challenges such as different individuals wearing similar clothing. Given its importance in real-world surveillance and security applications ~\cite{cornett2023expanding, Gupta_2024_WACV}, we focus on addressing CC-ReID in this work.

Existing CC-ReID approaches often use external biometric modalities such as gait~\cite{Jin_2022_CVPR}, face~\cite{Wan_2020_CVPR_Workshops}, and body shape~\cite{Chen_2021_CVPR}, but their \textit{pre-processing cost} and reliance on additional models limit real-world applicability. RGB-only methods like CAL~\cite{gu2022clothes} learn clothing-invariant features but require \textit{external clothing annotations}, adding manual overhead.  
Recent works~\cite{Liu_2024_CVPR, He_2024_CVPR} leverage \textit{fine-grained attributes} (\eg black pants, has umbrella), but these are \textit{computationally expensive}, \textit{short-term} (per frame), and sensitive to \textit{clothing modifications}, \textit{occlusion}, and \textit{illumination changes}, making them unreliable over time. 
For practical deployment, it is crucial to reduce reliance on such external attributes while effectively handling clothing variations.

\begin{figure*}[t]
\centering
\includegraphics[width=0.98\linewidth]{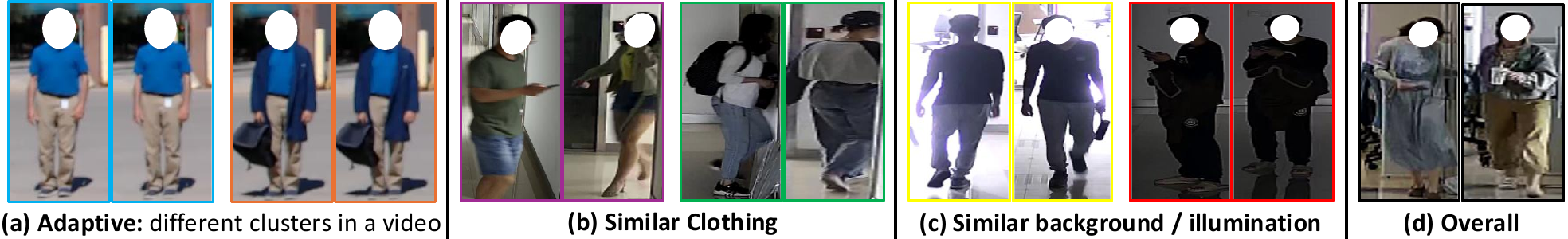} 
\vspace{-4pt}
\caption{\textbf{Clustering of color representation}: Clusters (2 images/cluster, same outline color) generated via RGB-uv projection~\cite{afifi2019SIIE}. \textit{(a):} Clothing changes in video are clustered differently \textit{(b \& c):} Clustering based on similar clothing \& background / illumination. \textit{(d):} Different clothing but similar overall color profile clustered together. Images from  CCVID (a) and LTCC (b, c, d).
}
\label{fig:raw_colors_clusters}
\vspace{-3.5pt}
\end{figure*}

Our work explores a simple, efficient color-based CC-ReID approach, eliminating the need for external annotations. 
By leveraging color cues, our method offers a computationally lightweight yet effective alternative for reducing appearance bias, particularly when clothing annotations are unavailable.
Color information in an image possesses three key properties:  
\textbf{1) Efficient}: colors can be extracted in real-time with minimal overhead  
\textbf{2) Adaptive}: colors dynamically adjust to changes in illumination, occlusion, and environmental conditions  
\textbf{3) Contextual}: colors capture both clothing and background information, enabling differentiation between similar clothing in different contexts.
\Cref{fig:raw_colors_clusters} illustrates the adaptive and contextual properties through clustering of images via RGB-derived color representations.

We propose \textbf{Color See, Color Ignore (CSCI)}, a transformer-based method that learns color representations (\textit{Color See}) from raw RGB-only input while disentangling them from ReID features (\textit{Color Ignore}).  
CSCI introduces a dedicated \textbf{Color token}, analogous to the class token in ViTs, to predict color embeddings, enabling explicit color representation learning. 
However, the \textit{parameter-efficient} nature of this token risks color bias leakage into ReID features via self-attention. 
To address this, we propose a middle-ground solution between two extremes: (1) complete overlap of ReID and Color (appearance) tokens~\cite{gu2022clothes, han2023clothing, Nguyen_2024_WACV}, and (2) fully separate dual-branch~\cite{Liu_2024_CVPR,    Yang_2023_CVPR} via our \textbf{S2A self-attention}, enabling 
ReID and Color tokens to perform independent self-attention, preventing cross-information transfer. 
We demonstrate CSCI’s effectiveness and generalization across both image and video ReID tasks.

In summary, 1) `CSCI' learns to ignore color representation using RGB-only inputs \textit{without} any external annotation or modality. 
2) Using the Color token and S2A self-attentions, we disentangle color bias from the ReID features.
3) CSCI outperforms existing methods on two image CC-ReID benchmarks: LTCC and PRCC, and two video benchmarks: CCVID and MeVID.
4) The ablation study provides evidence of colors' viability as a cost-efficient alternative for external annotations/modalities.

\section{Related Work}

\noindent \textbf{CC-ReID}
External modalities such as gait~\cite{Fan_2023_CVPR,Azad_2024_CVPR}, face~\cite{arkushin2024geff},
body-shape~\cite{Liu_2023_ICCV}, background~\cite{bmvc_lcccpr}, pose~\cite{Nguyen_2024_WACV} \etc. are often used to capture biometric signatures. 
However, these modalities have computationally expensive preprocessing, limiting their application. 
Conversely, RGB-only models~\cite{gu2022clothes, Yang_2023_CVPR, han2023clothing} are less resource-intensive but rely on external clothing annotations~\cite{wang2024exploring}.
Moreover, these attributes are usually `short-term', \ie clothing can change throughout a video, potentially requiring frame-level annotations.
To address these issues, we propose learning the RGB-only color representations, 
that can serve as cost-free frame-level annotations for appearance bias removal without relying on external annotations. 
In the past, colors have mostly been used as augmentation~\cite{shu2021semantic, Liu_2023_ICCV} rather than learnable attributes. 
Our approach can potentially offer a more cost-efficient CC-ReID in the absence of clothing attributes.  
\vspace{3pt}

\noindent\textbf{Transformers in ReID } TransReID~\cite{he2021transreid}, and follow-ups~\cite{Bansal_2022_WACV, chen2023beyond} have used ViT class token for ReID.
Other early works have combined traditional ReID with transformers, \eg memory banks~\cite{ni2023part},
external modalities~\cite{CHEN202290}, CNNs features~\cite{jia2022learning} \etc).
Recent methods~\cite{Liu_2024_CVPR, He_2024_CVPR, wang2024image, Liang_2025_CVPR,Azad_2025_ICCV} incorporate text-based fine-grained descriptions into the ReID features.
\citeauthor{ye2024transformer} surveyed the growing popularity of transformers in ReID, with state-of-the-art largely dominated by transformers.
With the growing complexity of transformers, the added cost of external attributes could become a bottleneck. 
Thus our design primarily focuses on transformer-centric solutions, via a computationally inexpensive learnable token to learn and disentangle color-related features.  

\begin{figure}[!t]
    \centering
    \includegraphics[width=0.98\linewidth]{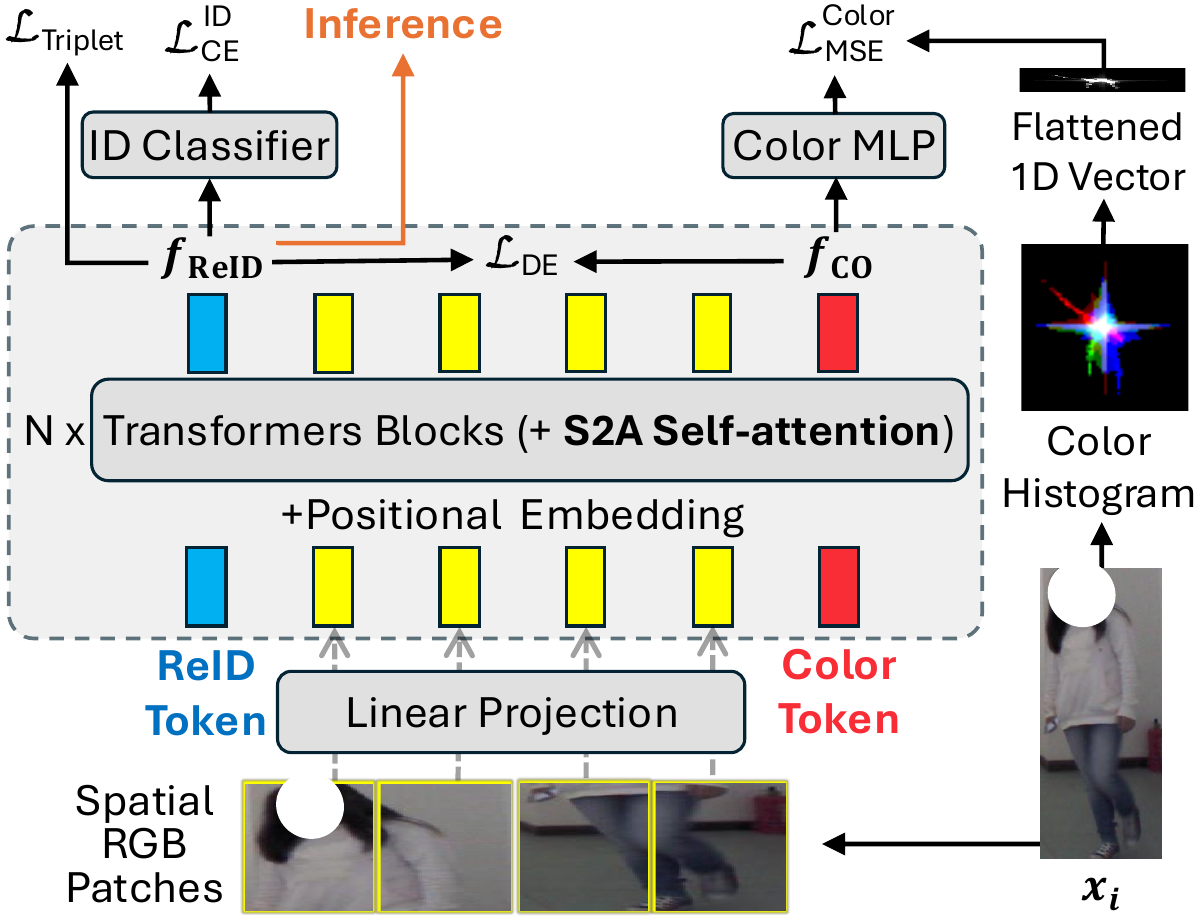}
    \vspace{-3pt}
    \caption{
    \textbf{CSCI Framework}: Image spatial tokens (yellow) are appended with ReID (\BLUE{$\boldsymbol{f_{ReID}}$}), and Color (\RED{$\boldsymbol{f_{CO}}$}) tokens. Tokens are passed through transformer blocks (modified self-attention), discarding spatial ones after the last block. $f_{ReID}$ (inference) predicts the identity labels and trains with triplet loss. $f_{CO}$ predicts flattened color histograms and disentangled from $f_{ReID}$. 
    }
    \label{fig:pipeline}
    \vspace{-3.5pt}
    \end{figure}
    
    \section{Methodology}

    \begin{figure*}[!th]
    \centering
    \begin{subfigure}{.25\textwidth}
      \centering  \includegraphics[height=4cm]{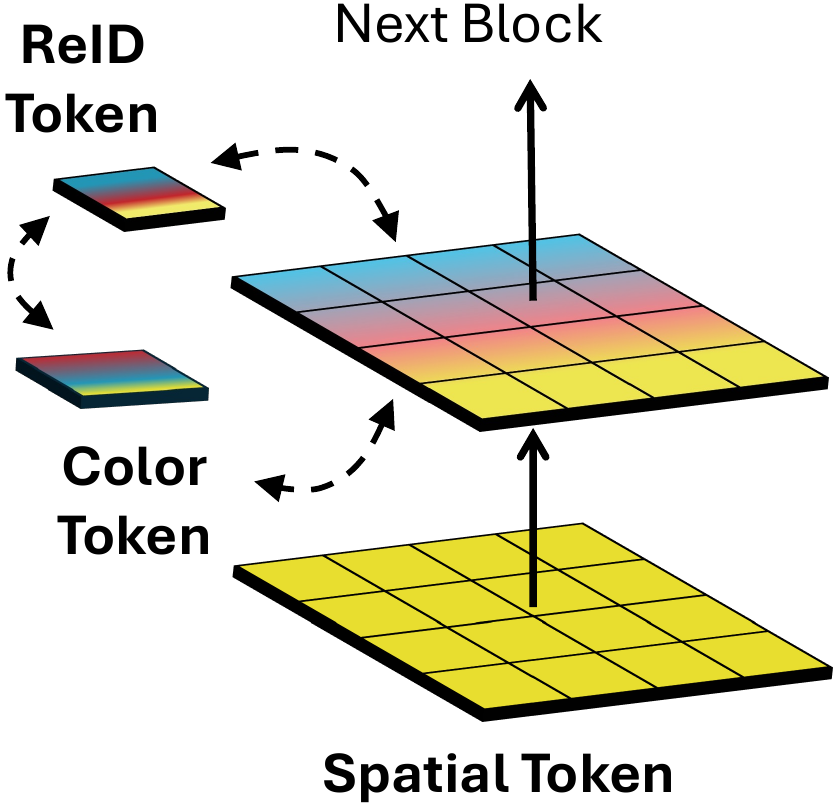}
      \caption{Tradational Self-Attention}
      \label{fig:SA1}
    \end{subfigure}
    \qquad
    \begin{subfigure}{.25\textwidth}
      \centering  \includegraphics[height=4cm]{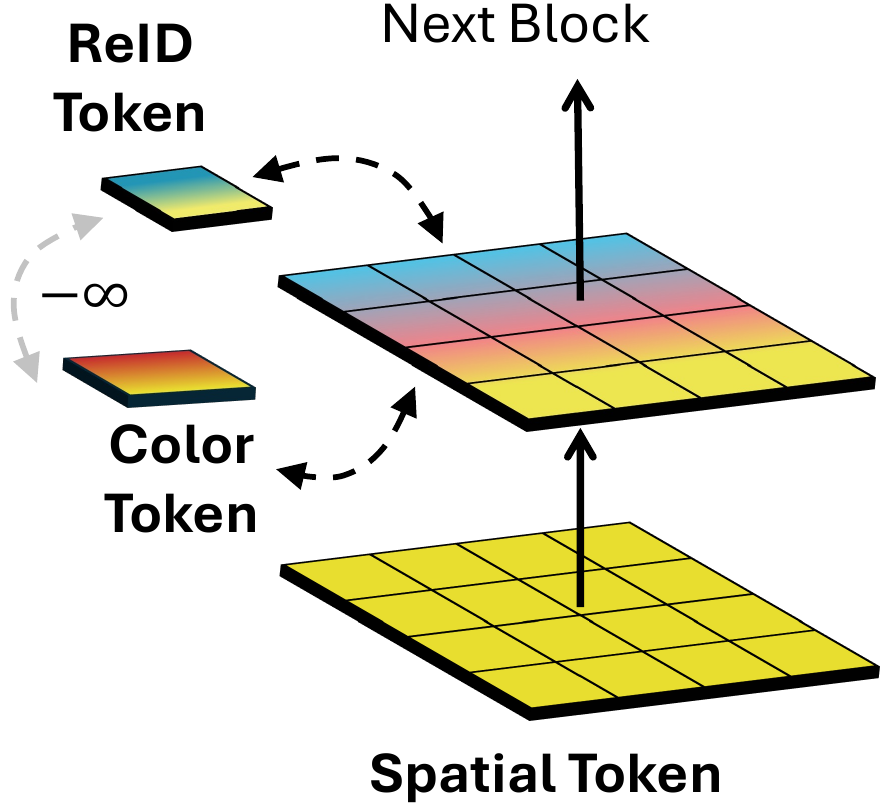}
      \caption{Masked Self-Attention}
      \label{fig:SA2}
    \end{subfigure}%
    \qquad
    \begin{subfigure}{.35\textwidth}
      \centering  \includegraphics[height=4cm]{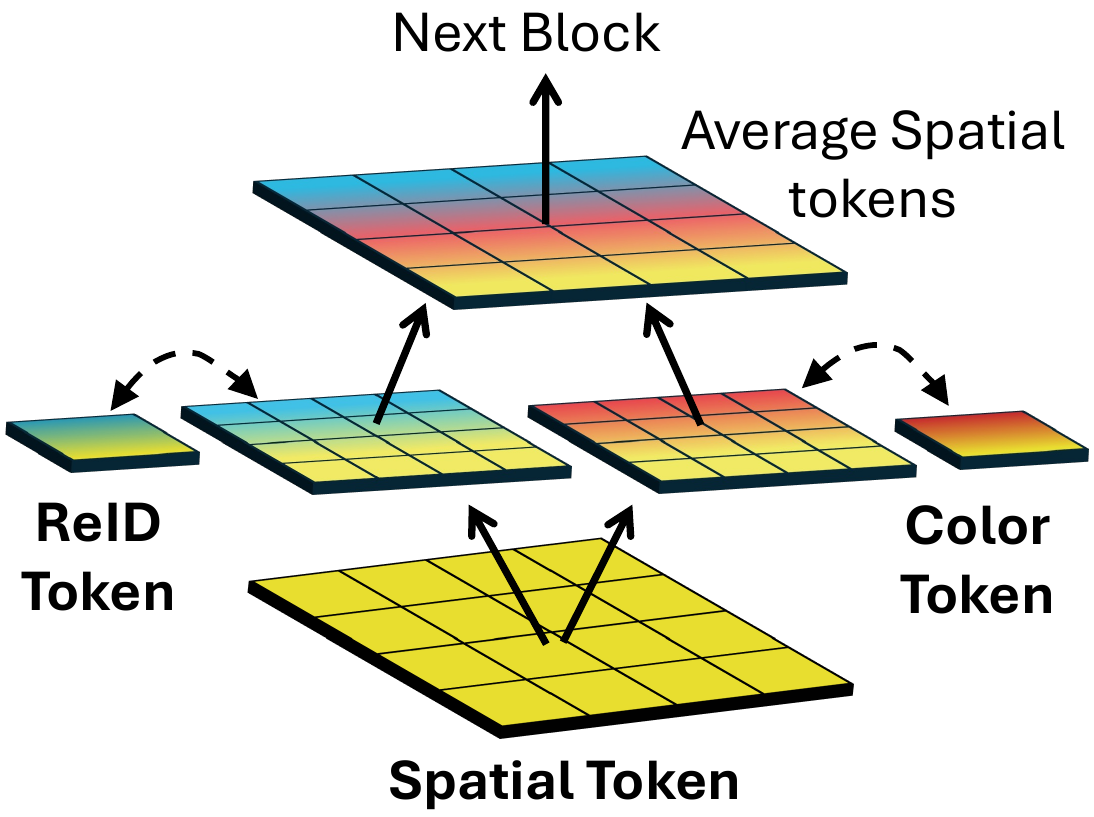}
      \caption{S2A Self-Attention}
      \label{fig:SA3}
    \end{subfigure}%
    \caption{
    \textbf{Self-Attention Variants}:
    Dotted arrows and shades of coloring indicate information exchange via self-attention.
    For explanation, spatial tokens are grouped as one, skipping their internal self-attention.
    \textit{(a) Traditional self-attention}, all tokens attend each other;
    \textit{(b) Masked self-attention} is the same as traditional without \textit{ReID-Color} token attention;
    \textit{(c) S2A self-attention} has two independent self-attention steps, one for ReID and the other for Color token. Spatial tokens are averaged across the two steps.
    }
    \label{fig:self-attn_variants}
    \end{figure*}

\textbf{Color See, Color Ignore (CSCI)}, learns to generate color-agnostic CC-ReID features in a parameter-efficient manner. 
Given a training set of \textit{N} data points $(x_i, y_i)_{i=1}^N$, where $x_i$ are RGB images and $y_i$ are identity labels, \textbf{CSCI does not rely on any external or internal (clothing) annotations}. During both training and testing, only RGB images $x_i$ are provided as input, producing ReID features $f_{REID}$ and color embeddings $f_{CO}$ (\cref{fig:pipeline}), while only $f_{REID}$ is used for inference. 
Inference involves matching query $f_{REID}$ against images/videos in the gallery.
A dedicated \textbf{Color token}  learns color embeddings, mitigating the color-induced appearance bias.
We employ a two-step self-attention mechanism, \textbf{S2A Self-Attention}, to prevent information leak between ReID and color features. 
In this section, we describe the each component of CSCI.

    \subsection{Color See Color Ignore (CSCI)}
    The idea behind our proposed method \textit{`Color See Color Ignore'} is to learn color representation from RGB-only inputs (\textit{`See Color'}) and using the learned color representation, we disentangle it from the ReID features (\textit{`Color Ignore'}).  An overview of the proposed method is shown in \Cref{fig:pipeline}.

    \subsubsection{Image Encoder}
    We propose a transformers-based approach where the input image is first converted into non-overlapping spatial tokens (yellow). 
    Two identical learnable head tokens, ReID token (popularly called `class token') (blue), and Color token (red) are appended to these spatial tokens. 
    All tokens along with their positional embedding are passed through a series of transformer blocks (w/ modified self-attention).
    ReID and Color tokens derive their respective embeddings from the context of spatial tokens.     
    
    After the last block, spatial tokens are discarded, ReID token serves as ReID features $f_{ReID}$, while the Color token represents color embeddings $f_{CO}$.
    An ID classifier head is added on top of $f_{REID}$ for classifying the identity labels $y_i$ of the individuals in the training set.
    On the other hand, $f_{CO}$ learns color embedding by predicting/regressing the color representation vector via a fully connected MLP layer. 
    Predicting color vectors removes test time dependency on these representations.
    The cosine distance between $f_{REID}$ and $f_{CO}$ is maximized (inducing orthogonality~\cite{li2023dc}) for removing color bias from ReID features.
    \begin{align}
        \mathcal{L}_{DE}(f_{CO},f_{REID}) = \bigg| \frac{f_{CO}}{\| f_{CO} \|_2} \cdot \frac{f_{REID}}{\| f_{REID} \|_2} \bigg| 
        \label{eq:dist_loss}
    \end{align}
    This additional head token-based approach~\cite{touvron2021training} is parameter-efficient and can help generate color embeddings in the same feature space as that of the ReID token. We adapt EVA-02~\cite{fang2024eva} vision transformer for image encoder in our approach. In addition, we also show its generalization to other existing transformer-based approaches for ReID.
    
    \subsubsection{S2A Self-Attention}
    In \textit{Traditional Self-Attention} (100\% overlap of color and ReID, \cref{fig:SA1}), the ReID (class) token attends (exchanges information) with every other token, and vice versa. 
    It's commonly depicted for one head self-attention without token dimension as, 
    $Att(Q,K,V) = softmax(Q.K^T)V$
    where Q, K, and V are Query, Key, and Value representations. The weights assigned to each token are shown as \vspace{-14pt}
    \begin{align}
    QK^T&=\begin{array}{@{\hspace{0.4em}}c@{\hspace{0.4em}}|@{\hspace{0.4em}}ccc@{}}
     & k_{CO} & k_{ID} & k_{SP} \\
    \hline
    q_{CO} & q_{CO} \cdot k_{CO} & q_{CO} \cdot k_{ID} & q_{CO} \cdot k_{SP} \\
    q_{ID} & q_{ID} \cdot k_{CO} & q_{ID} \cdot k_{ID} & q_{ID} \cdot k_{SP} \\
    q_{SP} & q_{SP} \cdot k_{CO} & q_{SP} \cdot k_{ID} & q_{SP} \cdot k_{SP} 
    \label{eq:tradation_sa_wt}
    \end{array}
    \end{align}
    Here, the Color token (CO) exchanges information directly with the ReID token (ID) and spatial token(s) (grouped as SP), risking entangling color bias into ReID features.  
    A simple solution is to use \textit{Masked Self-attention}~\cite{Devlin2019BERTPO}, where tokens are masked with $-\infty$ weights (0 after softmax, \cref{fig:SA2}). When computing self-attention for the ReID token, Color token is assigned $-\infty$ weight, and similarly Color token assigns $-\infty$ weight to the ReID token.\vspace{-5pt}
    \begin{align}
QK^T_{\text{Mask}}\hspace{0.1em}&=\hspace{0.1em}\begin{array}
    {@{}c@{\hspace{0.1cm}}|@{\hspace{0.1cm}}c@{\hspace{0.28cm}}c@{\hspace{0.28cm}}c@{\hspace{0.1cm}}}
     & k_{CO} & k_{ID} & k_{SP}\\
    \hline
    q_{CO} & q_{CO}\cdot k_{CO} & -\infty & q_{CO}\cdot k_{SP} \\
    q_{ID} & -\infty & q_{ID} \cdot k_{ID} & q_{ID}\cdot k_{SP} \\
    q_{SP} & q_{SP}\cdot k_{CO} & q_{SP}\cdot k_{ID} & q_{SP} \cdot k_{SP}
    \label{eq:masked_sa_wt}
    \end{array}
    \end{align}
    While this reduces the direct information leak between the Color and ReID tokens, spatial tokens however are exposed to both ReID and Color tokens. Both tokens influence each other’s weight (\cref{eq:masked_sa_wt} last row) on final spatial (context) embedding.
    We believe that the ReID token, representing biometrics, is as important as the appearance bias represented by the Color token. 
    Therefore, to give equal weight to both tokens, we separate the self-attention step entirely for ReID and Color tokens.
    We name this \textbf{\textit{S2A Self-Attention}} (\cref{fig:SA3}).
    Mathematically, it can be represented as two self-attention steps, one without the Color token ($\sim CO)$, \cref{eq:sep_sa_no_co}), and one without the ReID token ($\sim ID)$, \cref{eq:sep_sa_no_cl}), and averaging for Spatial Tokens \cref{eq:sep_sa_avg_sp}). \vspace{-12pt}
    \begin{align}
    QK^T_{\sim CO} &= \begin{array}{c|@{\hspace{0.5em}}cc@{}}
     & k_{ID} & k_{SP} \\
    \hline
    q_{ID} & q_{ID} \cdot k_{ID} & q_{ID} \cdot k_{SP} \\
    q_{SP} & q_{SP} \cdot k_{ID} & q_{SP} \cdot k_{SP}
    \label{eq:sep_sa_no_co}
    \end{array} \\
    QK^T_{\sim ID} &= \begin{array}{c|cc}
     & k_{CO} & k_{SP} \\
    \hline
    q_{CO} & q_{CO} \cdot k_{CO} & q_{CO} \cdot k_{SP} \\
    q_{SP} & q_{SP} \cdot k_{CO} & q_{SP} \cdot k_{SP}
    \label{eq:sep_sa_no_cl}
    \end{array}
    \end{align}
    \vspace{-11pt}
    \begin{align}
    Att(Q^{SP},K^{SP},V^{SP}) &= \Big( Att \big(Q_{\sim ID}^{SP},K_{\sim ID}^{SP},V_{\sim ID}^{SP} \big) + \nonumber \\ 
    & \hspace{-0.4cm} Att\big(Q_{\sim CO}^{SP},K_{\sim CO}^{SP},V_{\sim CO}^{SP}  \big) \Big) / 2
    \label{eq:sep_sa_avg_sp}
    \vspace{-15pt}
    \end{align}
    These two self-attention(s) add minimal computational overhead while outperforming its counterparts (\cref{tab:sa_design}).
An alternative would be to use dual-branch (one for color, and the other for ReID)~\cite{pathak2025coarse, Yang_2023_CVPR} to completely nullify the overlap of these tokens, \textit{computationally impractical}. 
S2A with some leak between ReID and color token offers a middle ground between complete (\cref{eq:tradation_sa_wt}) and no overlap. 

\subsection{Color Representation}
\begin{figure}[!ht]
    \centering
    \begin{subfigure}{.47\textwidth}
    \centering \includegraphics[width=0.9\linewidth]{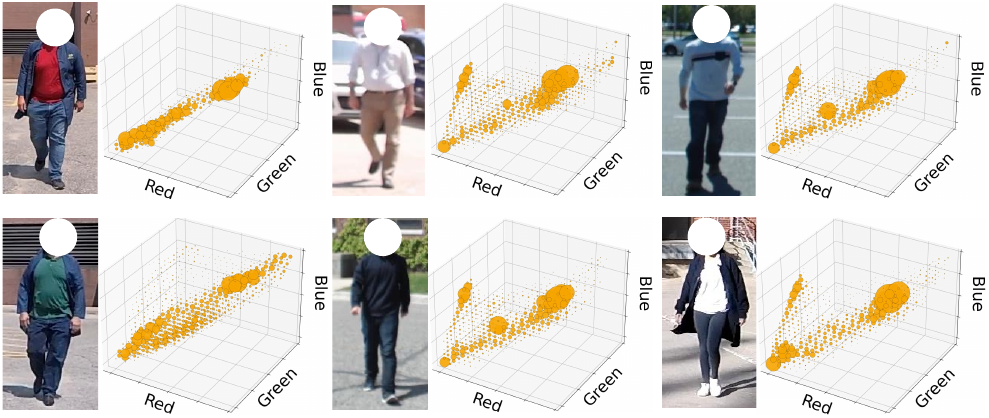}
    \label{fig:color_hist_pix}
    \end{subfigure}
    \qquad
    \begin{subfigure}{.47\textwidth}
    \centering \includegraphics[width=0.9\linewidth]{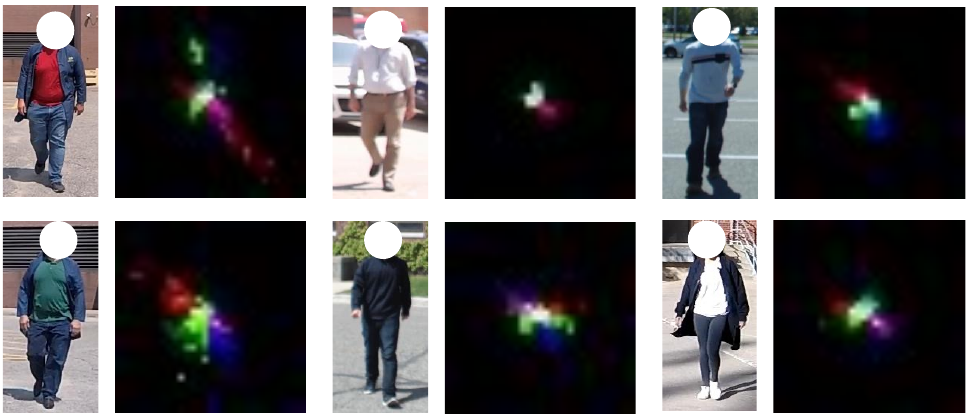}
    \label{fig:color_hist_rgb-uv}
    \end{subfigure}
    \caption{\textbf{Color Representation}: 
    \textit{Top Pixel Binning:}
     3D Histogram $(20\x20\x20)$; \textit{Bottom:RGB-uv 2D Projection/Histogram} $(3\x32\x32$, upscaled for viewability.
    Images taken from CCVID.
    }
    \label{fig:color_hist_viz} 
    \vspace{-4pt}
    \end{figure}

    Colors from images/video frames can provide an \textit{adaptive} (current state \eg illumination) and \textit{contextual} (clothes and background colors) profile of the input (\cref{fig:raw_colors_clusters}).
    Here, we explore two different color representations: \textit{`Pixel Binning (Pix Bin)'} and 
    \textit{`RGB-uv projection (RGB-uv)'}~\cite{afifi2019SIIE}.
    For a given bin size $h$, 
    \textit{Pix Bin} creates a 3D histogram based on pixel intensity/values (0-256) and counts the number of pixels in each bin. 
    This generates a $h\times h\times h$ 3D histogram.
    On the other hand, RGB-uv generates 2D histogram projections ($h\times h$) for each R,G,B channel, via a set of differential operations (\SUPP)\footnote{Code:  \url{https://github.com/mahmoudnafifi/HistoGAN}}.
    This generates a $3\times h\times h$ histogram.
    RGB-uv projections have been shown to effectively style transfer in GANs~\cite{afifi2021histogan}.
    Any of these methods can be used to represent colors (\cref{fig:color_hist_viz}), and by flattening these histograms into 1D vectors we can represent these histograms as ground truth color representations. 
    Both, `Pix Bin' and `RGB-uv' can be computed efficiently on the fly from RGB images \ie without any external dependency.

\begin{table*}[!t]
\centering
\renewcommand{\arraystretch}{1.05}
\setlength\tabcolsep{5pt}
\scalebox{0.9}{

    \begin{tabular}{l|c| c|c|c| cc| cc|| cc| cc}
    \specialrule{1pt}{0pt}{0pt}
    \rowcolor{mygray} 
    \multicolumn{2}{c|}{} & & \multicolumn{2}{c|}{Additional}
    & \multicolumn{4}{c||}{LTCC} & \multicolumn{4}{c}{PRCC} \\
    [-.4pt]\hhline{~~~~~--------}
    \rowcolor{mygray} 
    \multicolumn{2}{c|}{} & & \multicolumn{2}{c|}{Attributes}
    & \multicolumn{2}{c|}{CC}  & \multicolumn{2}{c||}{General} & \multicolumn{2}{c|}{CC} & \multicolumn{2}{c}{SC} \\ 
    [-.4pt]\hhline{~~~----------}
    \rowcolor{mygray} 
    \multicolumn{2}{c|}{\multirow{-3}{*}{Method}} & \multirow{-3}{*}{Venue} & Int. & External & R-1$\uparrow$ & mAP$\uparrow$ & R-1$\uparrow$ & mAP$\uparrow$ & R-1$\uparrow$ & mAP$\uparrow$ & R-1$\uparrow$ & mAP$\uparrow$ \\
    \hline\hline

    \multirow{3}{*}{\rotatebox[]{90}{ResNet}} 
    & CAL \cite{gu2022clothes} & CVPR'22 & CL & - & 
       40.1 & 18.0 & 74.2 & 40.8 & 55.2 & 55.8 & \B{100} & \SB{99.8}\\ 
    &    CCFA~\cite{han2023clothing} & CVPR'23 & CL & - & 45.3 & 22.1 &  75.8 & 42.5 &  61.2 & 58.4 & 99.6 & 98.7 \\ 
    & FIRe$^2$~\cite{wang2024exploring} & TIFS'24 & - & - & 44.6  & 19.1 & 75.9 & 39.9 &  65.0 & 63.1 & \B{100}  & 99.5  \\ 

       \hline 
    \multirow{5}{*}{\rotatebox[]{90}{Multi-ResNets}}  &  3DInvar.~\cite{Liu_2023_ICCV}& ICCV'23 
        & CL & Po+BS & 40.9 & 18.9 
        & - & - & 56.5 & 57.2 & - & - \\ 
    &    AIM \cite{Yang_2023_CVPR} & CVPR'23 & CL & - & 40.6 & 19.1 & 76.3 & 41.1 & 57.9 & 58.3 & \B{100} & \B{99.9} \\
     
    &    DCR.
        \cite{10036012}& TCSVT'23
        & CL & BP+Co. & 
        41.1 & 20.4 & 76.1 & 42.3 & 57.2 &  57.4 & \B{100} & 99.7 \\ 

    
    & CVSL~\cite{Nguyen_2024_WACV}& WACV'24 & CL & Po &  44.5 & 21.3 & 76.4 &  41.9 & 57.5 & 56.9 & 97.5 & 99.1 \\ 
    
    & CCPG~\cite{Nguyen_2024_CVPR} & CVPR'24 &CL & Po& 46.2 & 22.9 & 77.2 & 42.9 & 61.8 & 58.3 & \B{100} & 99.6\\ 

    \hline 
    \cellcolor{white}& CLIP3D~\cite{Liu_2024_CVPR} & CVPR'24 & - &BS+Text & 42.1 &  21.7 & - & - &  60.6 &  59.3 &- & - \\
    
    \cellcolor{white} & IRM~\cite{He_2024_CVPR}$\dagger$ & CVPR'24 & CL & Text+BP& \SB{46.7} & - & 66.7 & - & - & - & - & - \\ 
    [-.4pt]\hhline{~------------}
    \cellcolor{white} & \multicolumn{2}{l|}{EVA-02} 
     & - & - & 44.9	& 23.1 & 80.3 & 45.9 & 61.6 & 59.0 & \B{100} & \B{99.9} \\
     [-.4pt]\hhline{~------------}

     \rowcolor{mygray}
     \cellcolor{white} &  \multicolumn{2}{l|}{CSCI - Pix. Bin \textit{(our)}} & - & - & 			
    \SB{46.7}\IMPROV{+1.8} & \SB{23.6}\IMPROV{+0.5} & 80.8\IMPROV{+0.5}  & 45.9  & \B{66.6}\IMPROV{+5.0} & \SB{60.7}\IMPROV{+1.7} & \B{100} & \B{99.9} \\ 

    \rowcolor{mygray} 
    \multirow{-5}{*}{\rotatebox[]{90}{Transformer}} 
    \cellcolor{white} 
    &  
    \multicolumn{2}{l|}{CSCI - RGB-uv \textit{(our)}} 
    & - & - & 
    \B{47.8}\IMPROV{+2.9} & \B{24.4}\IMPROV{+1.3} & \SB{82.6}\IMPROV{+2.3} & \SB{48.0}\IMPROV{+2.1} & \SB{66.2}\IMPROV{+4.6} & \B{61.3}\IMPROV{+2.3} & \B{100} & \B{99.9} \\  
    
    \specialrule{1pt}{0pt}{0pt}
    \end{tabular}
    }
    \vspace{-5pt}
    \caption{
        \textbf{LTCC \& PRCC Results (\%)}. 
        Shorthand notations are: \B{Best results} and \SB{second best results},  
        `Int.': internal (`CL' clothing labels), `Po': Pose/skeleton, `Co': Contours, `BS': 3D Body shape, `BP': Body parse silhouettes,
        `-': results/attributes not reported/used, 
        $\uparrow$: Higher means better performance,
        $\dagger$, IRM multi-dataset training results skipped. 
        \textbf{The improvement X\% over the baseline is indicated by \IMPROV{+X}
    }.
}
\label{tab:ltcc_prcc_results}
\vspace{-4pt}
\end{table*}

    \begin{table*}[!th]
    \centering
\renewcommand{\arraystretch}{1.1}
    \setlength\tabcolsep{4pt}
    \scalebox{0.9}{
    \begin{tabular}{c|c|c| c|c| cc|cc|| cccc}
    \specialrule{1pt}{0pt}{0pt}
    \rowcolor{mygray} 
    \multicolumn{2}{c|}{}& & & & \multicolumn{4}{c||}{CCVID} & \multicolumn{4}{c}{MeVID} \\
    [-.4pt]\hhline{~~~~~--------}
    \rowcolor{mygray} 
    \multicolumn{2}{c|}{}  & & \multicolumn{2}{c|}{\multirow{-2}{*}{\begin{tabular}[c]{@{}c@{}} Additional \\ Attributes\end{tabular}}} & \multicolumn{2}{c|}{CC} & \multicolumn{2}{c||}{General} & \multicolumn{4}{c}{Overall (General)} \\ 
    [-.4pt]\hhline{~~~----------}
    \rowcolor{mygray} 
    \multicolumn{2}{c|}{\multirow{-3}{*}{Method}}
     & \multirow{-3}{*}{Venue} & Int. & External & R-1$\uparrow$ & mAP$\uparrow$ & R-1$\uparrow$ & mAP$\uparrow$ & R-1$\uparrow$ & R-5$\uparrow$ & R-10$\uparrow$ & mAP$\uparrow$ \\
    \hline \hline
    & CAL 
       \cite{gu2022clothes} & CVPR'22 & CL & - & 81.7 & 79.6 & 82.6 & 81.3 & 52.5 & 66.5 & 73.7 & 27.1 \\ 
       \hline
       & 3DInvar. 
        \cite{Liu_2023_ICCV} & ICCV'23
        & CL & Po+BS & 84.3 & 81.3 & 83.9 & 82.6 & - & - & - & - \\ 
       & SEMI 
        \cite{nguyen2024temporal} & WACV'24 
        & - & BS & 82.5 & 81.9 & 83.1 & 81.8 & - & - & - & -  \\ 
        \multirow{-3}{*}{\rotatebox[]{90}{M-CNN}} &  ShARc
        ~\cite{Zhu_2024_WACV}& WACV'24
        & - & Sil+BS+Po & 84.7 & 85.2 & 89.8 & 90.2 & 59.5 & 70.3 & 77.2 & 29.6 \\
    
    \hline
        & CLIP3D
        ~\cite{Liu_2024_CVPR} & CVPR'24
        & & BS+Text & 82.4 & 83.2 & 84.5 & 83.9 & - & - & - & -  \\ 
        & GBO
        ~\cite{li2024rethinking} & ArXiv'24 
        & CL & BP+Po+Diff & 86.9 & 83.5 & 89.7 & 87.1 & - & - & - & -  \\ 
        [-.4pt]\hhline{~------------}
        \cellcolor{white} & \multicolumn{2}{l|}{EVA-02 Image} & - & - & 
    86.4 & 87.0 & 88.8 & 89.0 &
    73.5 & 83.7	& 86.2 & 47.5 \\
    
& \multicolumn{2}{l|}{EVA-02 + EZ-CLIP~\cite{ahmad2023ez} } & - & - & 
89.8 & 90.1 & 91.1 & 91.1 & 
\SB{76.6} & 84.8 & 87.0 & 55.5 \\
[-.4pt]\hhline{~------------}
\rowcolor{mygray} 
\cellcolor{white} & \multicolumn{2}{l|}{CSCI-V - Pix. Bin \textit{(our)}} &- & - & 
\SB{90.5}\IMPROV{+0.7} & \SB{90.8}\IMPROV{+0.7} & \SB{91.5}\IMPROV{+0.4} & \SB{91.6}\IMPROV{+0.5} &

\B{79.1}\IMPROV{+2.5} & \B{87.5}\IMPROV{+2.7} & 
\B{89.1}\IMPROV{+2.1} & 
\B{56.9}\IMPROV{+1.4} \\ 

\rowcolor{mygray} 
\multirow{-6}{*}{\rotatebox[]{90}{Transformer}}  \cellcolor{white} &  \multicolumn{2}{l|}{CSCI-V - RGB-uv \textit{(our)}} & - & - & \B{90.8}\IMPROV{+1.0} & \B{91.3}\IMPROV{+1.2} & \B{91.7}\IMPROV{+0.6} & \B{92.2}\IMPROV{+1.1} & 
    
    \B{79.1}\IMPROV{+2.5} & \SB{87.2}\IMPROV{+2.4} & \SB{88.6}\IMPROV{+1.6} & \SB{55.6}\IMPROV{+0.1}\\
    \specialrule{1pt}{0pt}{0pt}
    \end{tabular}
    }
    \vspace{-5pt}
    \caption{\textbf{CCVID \& MEVID Results (\%)}. 
    Same shorthand notation as \cref{tab:ltcc_prcc_results}. Additional attributes include `Diff.' for the Diffusion model and `Sil' for silhouettes.
    The baseline here is the Video ReID `EVA-02 + EZ-CLIP'. M-CNN refers to Multi-CNN architecture.}
    \label{tab:ccvid_mevid}
    \vspace{-8pt}
    \end{table*}

\subsection{Learning Objective}
EVA-02 ($f_{REID}$) is fine-tuned on target ReID like traditional ReID ~\cite{he2021transreid}, with standard cross-entropy loss ($\mathcal{L}_{CE}^{ID}$) on ID logits for predicting the identities of individuals in the training set, and Triplet loss ($\mathcal{L}_{Tripelt})$ on $f_{ReID}$.
    For predicting the color representation vector, we use MSE regression loss ($\mathcal{L}_{MSE}^{Color}$) on $f_{CO}$. 
    $\mathcal{L}_{DE}$ (\cref{eq:dist_loss}) disentangles $f_{ReID}$ with $f_{CO}$.
    Our overall training objective $L_{ReID}$ is defined as 
    \begin{align}
    L_{ReID}=\mathcal{L}_{CE}^{ID}+\mathcal{L}_{MSE}^{Color}+ \mathcal{L}_{Triplet}+\mathcal{L}_{DE}
    \end{align}
    Details of loss functions are in \SUPP.
    
    \subsection{Generalization to Video ReID}

We also demonstrate the generalization of CSCI to Video ReID (CSCI-V) by integrating our Color token with a lightweight video adaptation strategy. Specifically, we employ the EZ-CLIP~\cite{ahmad2023ez, ahmad2025tl} technique, which efficiently extends image-based transformers to videos using temporal tokens (detailed architecture in \SUPP).  
Following its design, we first pre-train the image model as a standard Image ReID model (without the Color token) using randomly sampled frames from video clips. We then fine-tune the model for video sequences by incorporating temporal tokens while training in a \textbf{two-step alternating process}:  
\textbf{1) Vanilla Video ReID}: The model processes video frames using temporal tokens, similar to standard Video ReID, without the Color token.  
\textbf{2) CSCI Image ReID}: The temporal tokens are disabled, effectively converting the model back into an image-based transformer. The Color token is then used to disentangle color information, training on randomly sampled video frames.  
\textit{Note:} Temporal and Color tokens are not trained simultaneously, as temporal tokens have their own self-attention mechanism. Combining this with our two-step S2A self-attention would introduce multiple levels of self-attention in each transformer block, significantly increasing computational overhead.

\begin{table}[!t]
\centering
\renewcommand{\arraystretch}{0.95}
\setlength\tabcolsep{2pt}
\scalebox{0.9}{
\begin{tabular}{l| c|c|c|c| cc}
\specialrule{1pt}{0pt}{0pt}
\rowcolor{mygray} 
& & & \multicolumn{2}{c|}{Attributes} & \multicolumn{2}{c}{LTCC (CC) } \\
[-.4pt]\hhline{~~~----}
\rowcolor{mygray} 
\multirow{-2}{*}{Model} & \multirow{-2}{*}{Venue} & \multirow{-2}{*}{\makecell{ViT\\Backbone}} & Int. & Ext. & R-1$\uparrow$ & mAP$\uparrow$ \\
\hline \hline
TransReID\cite{he2021transreid} & ICCV'22 & & - & - & 31.1 & 15.1 \\ 
\rowcolor{mygray} 
+CSCI & \textit{(ours)} & \multirow{-2}{*}{\shortstack{ViT-B\\(86 M)}} & - & - & 36.0 & 15.5 \\ 
\hline
TMGF~\cite{Li_2023_WACV} & WACV'23 &  & - & - & 32.9 & 17.5 \\ 
\rowcolor{mygray} 
+CSCI & \textit{(ours)} & \multirow{-2}{*}{\shortstack{ViT-S\\(22 M)}} & - & - & 39.5 & 20.9 \\ 
\hline
PAT~\cite{ni2023part} & ICCV'23 & & - & - & 31.1 & 13.9 \\ 
\rowcolor{mygray} 
+CSCI & \textit{(ours)} & \multirow{-2}{*}{\shortstack{ViT-B\\(86 M)}} & - & - & 32.7 & 13.8 \\ 
\hline
 & & & - & Pose & 35.2 & 19.0 \\ 
\multirow{-2}{*}{TCiP~\cite{wang2023transformer}} & \multirow{-2}{*}{CGI'22} & & CL & Pose & 36.5 & 19.0 \\ 
\rowcolor{mygray} 
+CSCI & \textit{(ours)} & 
\multirow{-3}{*}{\shortstack{ViT-B\\(86 M)}}
& - & Pose & 38.8 & 19.7 \\ 
\hline 
\hline 
 & & & - & & 44.9 & 23.6 \\ 
\multirow{-2}{*}{Baseline} & & 
 & CL & & 46.3 & 24.0 \\ 
\rowcolor{mygray} 
+CSCI & \textit{(ours)} & \multirow{-3}{*}{\shortstack{EVA-02\\(303 M)}} & - &  & 47.8 & 24.4 \\ 
\hline
\specialrule{1pt}{0pt}{0pt}
\end{tabular}
}
\vspace{-5pt}
\caption{\textbf{CSCI on other Transformers}. Only TCiP is CC-ReID}
\label{tab:other_models}
\vspace{-13pt}
\end{table}
\section{Experiments}

\subsection{Experiment Settings}
\noindent \textbf{Dataset}
To show the effectiveness of our method we use 2 \textit{Image CC-ReID} datasets: 
(i) \textbf{LTCC}~\cite{qian2020long}: $77$ training identities (IDs) (9,576 images) \& $75$ test IDs (7,543 images). 
(ii) \textbf{PRCC}~\cite{yang2019person}: $150$ training IDs (17,896 images) \& $71$ test IDs (10,800 images).    
For the \textit{video CC-ReID} dataset, we choose (iii) \textbf{CCVID}~\cite{gu2022clothes} has $75$ training IDs (948 clips) with $151$ test IDs (1908 clips) and (iv) \textbf{MeVID}~\cite{Davila2023mevid} has $104$ training IDs (6338 clips) with $54$ test IDs (1754 clips). 

\noindent \textbf{Evaluation}
We report Cumulative Matching Characteristics (CMC, rank 1(R-1) and rank 5 (R-5), \etc.) and Mean Average Precision (mAP) as an average of two runs. 
Evaluation for each dataset: 
LTCC and CCVID both use the Cloth-Changing (CC) (Query-Gallery have different clothes) and the General Protocol (entire test set).
MeVID only uses the General Protocol. Other protocols for MeVID in \SUPP.   
PRCC uses the CC and the Same-Clothes (SC) protocol (Query-Gallery wears the same clothes).

\begin{table*}[!t]
\centering
\subcaptionbox{\textbf{Self-Attention Variations}: `M' means in million, `G' is giga FLOPs.\label{tab:sa_design}}
{
\renewcommand{\arraystretch}{1}
\setlength\tabcolsep{3pt}
\scalebox{0.92}{
\begin{tabular}{l| c|c|  cc| cc}
\specialrule{1pt}{0pt}{0pt}
\rowcolor{mygray}
& &  &
\multicolumn{2}{c|}{LTCC} &
\multicolumn{2}{c}{PRCC} \\ 
[-.4pt]\hhline{~~~----}
\rowcolor{mygray}
\multirow{-2}{*}{\makecell{Method\\(Self-Attention)}} &\multirow{-2}{*}{\makecell{\# Param $\downarrow$ \\(M)}}  & \multirow{-2}{*}{ \makecell{\#FLOP$\downarrow$\\(G)}} & R-1 & mAP & R-1  & mAP \\
\hline\hline
Baseline  & 303.47 & 81.18 & 44.9 & 23.1 & 61.6 & 59.0 \\ 
Traditional \comm{(\cref{fig:SA1})} & 311.87 & 81.52\comm{\textsuperscript{+0.42\%}} & 46.8 & 23.5 & 63.5 & 60.3 \\ 
Masked \comm{(\cref{fig:SA2})} & 311.87 & 81.52\comm{\textsuperscript{+0.42\%}} & 46.7 & 23.7 & 65.3 & 61.3\\
\hline 
S2A \textit{(our)} \comm{(\cref{fig:SA3})} & 311.87 & 84.74\comm{\textsuperscript{+4.39\%}} & \textbf{47.8} & \textbf{24.4} & \textbf{66.2} & \textbf{61.3}\\
\specialrule{1pt}{0pt}{0pt}
\end{tabular}
\vspace{-5pt}
}
}
\hfill
\subcaptionbox{\textbf{Attributes Variation}.
\label{tab:att_variant}}
{
\renewcommand{\arraystretch}{1}
\setlength\tabcolsep{3pt}
\scalebox{0.92}{
\begin{tabular}{l|cc}
\specialrule{1pt}{0pt}{0pt}
\rowcolor{mygray}
& \multicolumn{2}{c}{LTCC}  \\ 
[-.4pt]\hhline{~--}
\rowcolor{mygray}
\multirow{-2}{*}{Method} & R-1 & mAP  \\
\hline\hline
Baseline& 44.9 & 23.1  \\ 
W/ Clothes & 46.3 & 24.0  \\
Grey Aug. & 38.3 & 18.5 \\
\hline 
Colors \textit{(Our)}& \textbf{47.8} & \textbf{24.4}\\
\specialrule{1pt}{0pt}{0pt}
\end{tabular}
\vspace{-5pt}
}
}
\hfill
\subcaptionbox{\textbf{Colors Design Variation}.\label{tab:color_design1}}
{
\renewcommand{\arraystretch}{1}
\setlength\tabcolsep{3pt}
\scalebox{0.92}{
\begin{tabular}{l|cc}
\specialrule{1pt}{0pt}{0pt}
\rowcolor{mygray}
 &
\multicolumn{2}{c}{PRCC} \\ 
[-.4pt]\hhline{~--}
\rowcolor{mygray}
\multirow{-2}{*}{Method} & R-1  & mAP \\
\hline\hline
Baseline & 61.6 & 59.0  \\ 
Feed & 63.1 & 57.6\\
Pred. (Cos) & 65.0 & 60.9 \\
\hline 
Pred. (MSE) & \textbf{66.2} & \textbf{61.3}\\
\specialrule{1pt}{0pt}{0pt}
\end{tabular}
\vspace{-5pt}
}
}
\vspace{-5pt}
\caption{\textbf{Design Choices}: All evaluations done on cloth changing (CC) protocol. \# of Params \& GFLOPs calculated for PRCC dataset with RGB-uv color $32\times 32=1024$-dim projections (averaged across channels).
}
\vspace{-3pt}
\end{table*}

\subsection{Implementation}

The input resolution of all input is $224 \times 224$ with batch size 8 and we use n=4 number of frames for videos. 
We use ViT-L backbone of EVA-02 with its official pre-trained weights ~\cite{dosovitskiy2020image,fang2024eva}, architectural details in \SUPP. 
Generalization to other transformers is in 
\Cref{sec:generic_transformers} (\cref{tab:other_models}). The image encoder uses absolute position embeddings, with 24 blocks, patch size 14, 16 multi-head self-attention and 1024 dimensional tokens.
Color and ReID tokens (and their positional embeddings) are initialized with pre-trained class token weights. 
The image and video models are trained on two 16GB and 32GB Tesla V100 GPUs, respectively. 
Pixel Binning is computed using OpenCV `cv2.calcHist' method, with bin size 20, creating a flattened $20\times20\times20=8000$-Dim color vector. For RGB-uv projections, hyperparameters were determined empirically (\textit{Supplementary} for all color-related hyperparameters). 
RGB-uv projections are averaged across 3 channels, except for LTCC where it is concatenated. Images and videos use bin sizes of 32 and 16.

\subsection{Comparison with SOTA Methods}
Additional attributes used by previous approaches are either \textit{`internal'} like clothing labels (w/ dataset) or \textit{`external'}, generated from off-the-shelf models. 
We have skipped the results for 
model agnostic technique-based methods like ReFace~\cite{arkushin2024geff},  CCVReID~\cite{wang2022benchmark}, and DCGN~\cite{10.1145/3643490.3661806} which improve the accuracy of existing methods with facial features.
\textit{Supplementary} contains all results. \vspace{5pt}

\noindent \textbf{Image ReID (LTCC \& PRCC)}
\Cref{tab:ltcc_prcc_results} shows CSCI with color disengagement outperforms existing methods on both LTCC and PRCC despite not using any additional attributes.
FIRe$^2$~\cite{wang2024exploring} is RGB-only approach (unsupervised clustering for pseudo attributes) with comparable performance on PRCC.
CNN models (w/ additional attributes) mostly fall short of EVA-02 baseline, reinforcing transformers superiority.
Other transformer-based approaches (CLIP3D~\cite{Liu_2024_CVPR},  IRM~\cite{He_2024_CVPR}) using large-scale Image-Text models, similar to our EVA-02.
Color disentanglement improvement over image baseline (Top-1) using   
RGB-uv projection (\textit{2.9\%-2.3\% on LTCC} and 4.6\% on PRCC) outperforms pixel binning (1.8\%-0.5\% on LTCC and \textit{5\% on PRCC}). \vspace{5pt}

\noindent \textbf{Video ReID (CCVID \& MeVID)}
\Cref{tab:ccvid_mevid} shows video adaptation of image baseline using EZ-CLIP significantly improves the performance of the vanilla Image ReID model.
Video adaptation with color disentanglement (CSCI-V) outperforms existing methods for CCVID and MeVID. 
For videos, RGB-uv projection (Top-1 gain with \textit{1.0\%-0.6\%} for CCVID and 2.5\% on MeVID) in general has the same improvement over baseline (in all categories) as that of pixel binning (Top-1 gain with \textit{0.7\%-0.4\%} for CCVID and \textit{2.5\%} for MeVID.
The EVA-02 baseline significantly outperforms prior MeVID results, likely due to strong alignment between real-world MeVID data and the model’s pre-trained weights.
Other foundation model-based GBO~\cite{li2024rethinking} uses diffusion to generate data augmentations, in addition to other attributes like pose, clothes, and body parsing; it has comparable results to ours.   
\textit{Efficient video adaptation using EZ-CLIP~\cite{ahmad2023ez, ahmad2025tl} incorporates image ReID achievements into video ReID, offering a viable future research direction.}   \vspace{5pt}

\noindent \textbf{Generalization of CSCI}
\label{sec:generic_transformers}
To demonstrate the generalization and fairness of our method, CSCI (Color token and S2A self-attention), we applied colors to other transformer backbones, specifically ViT-small (TMGF~\cite{Li_2023_WACV}, 22M parameters) and ViT-base (TransReID~\cite{he2021transreid}, TCiP~\cite{wang2023transformer}, and PAT~\cite{ni2023part}, 86M parameters), as shown in ~\underline{\Cref{tab:other_models}}. 
These backbones are significantly lighter than our backbone, which primarily uses a ViT-Large.
TransReID, PAT, and TMGF are the same clothes ReID models; hence, CSCI mainly reflects performance enhancement, particularly for TMGF, with a 7\% Top-1 improvement.
For the CC-ReID model TCiP, color outperforms clothing labels.

\begin{table}[!t]
\centering
\renewcommand{\arraystretch}{1}
\setlength\tabcolsep{3pt}
\scalebox{0.92}{
\begin{tabular}{l| cc}
\specialrule{1pt}{0pt}{0pt}
\rowcolor{mygray} 
Method & R-1$\uparrow$ & mAP$\uparrow$ \\
\hline \hline
Baseline & 44.9 & 23.1 \\ 
(Directly) Colors Vectors (1024-dim) & 45.7 & 21.4\\
(Directly) Project ReID to Colors (3072-dim) & 44.9 & 23.8\\
\rowcolor{mygray} 
Using Color tokens (3072-dim.) (Our) & 47.8 & 24.4 \\ 
\specialrule{1pt}{0pt}{0pt}
\end{tabular}
}
\vspace{-5pt}
\caption{\textbf{Learning vs directly using colors}. LTCC results.}
\label{tab:color_token_need}
\vspace{-5pt}
\end{table}

\begin{figure*}[!t]
\centering
\begin{subfigure}{.48\textwidth}
  \centering
\includegraphics[width=0.98\linewidth]{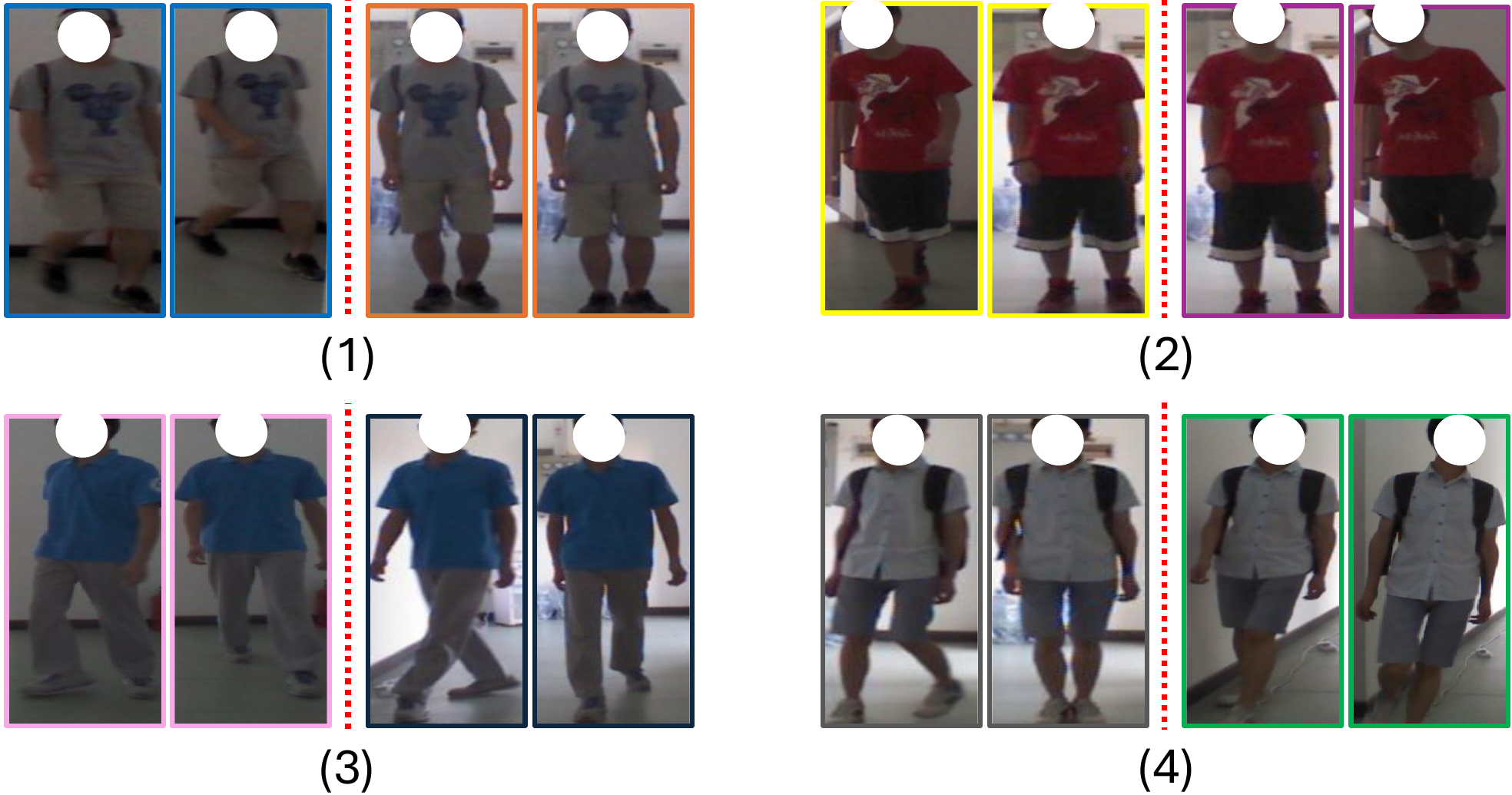}
\caption{\textbf{Same} cloth label \textbf{Different} k-means cluster of Color token}
\label{fig:SCDK}
\end{subfigure}%
\hfill
\begin{subfigure}{.48\textwidth}
  \centering
\includegraphics[width=0.98\linewidth]{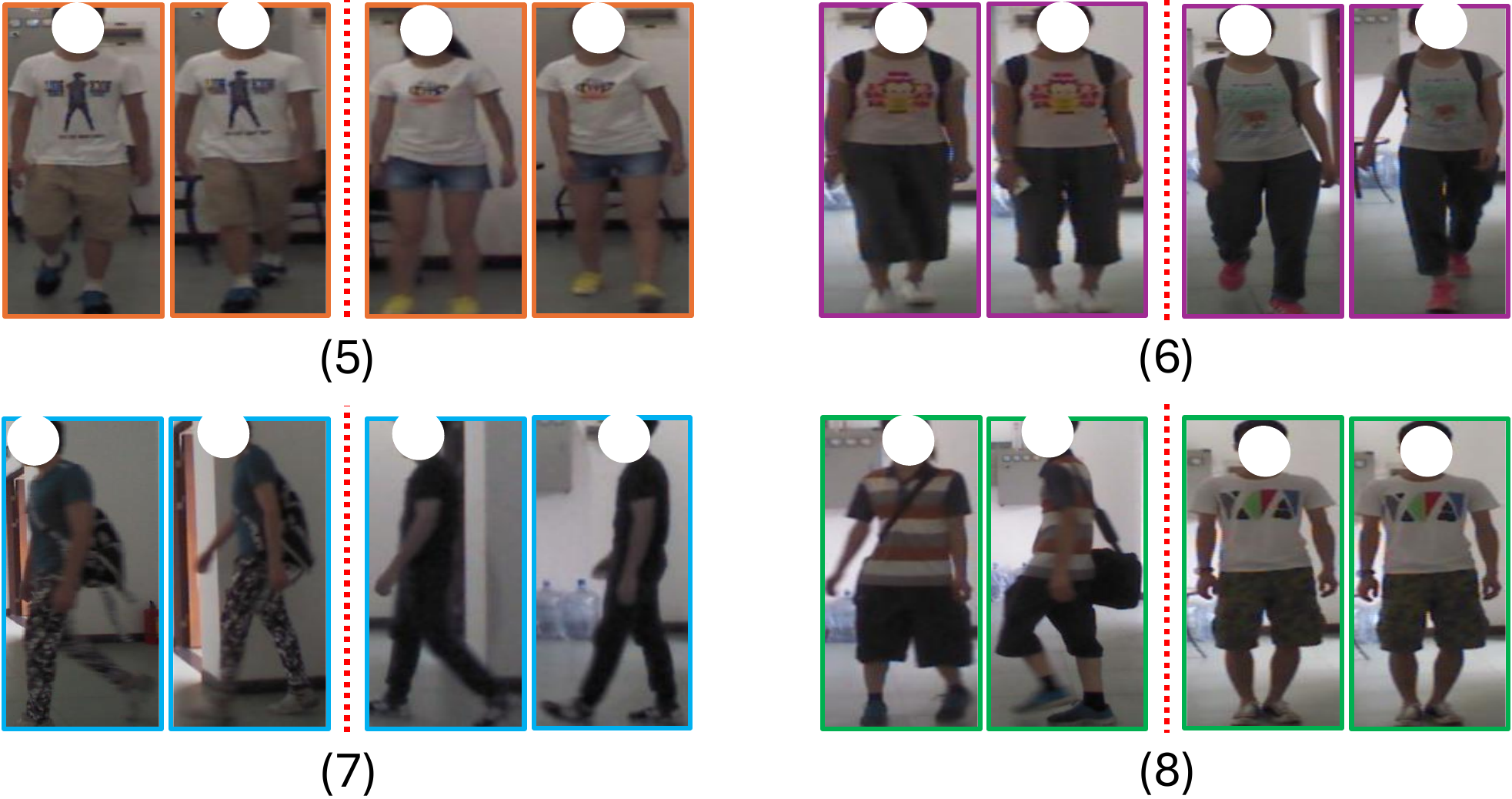}
\caption{\textbf{Different} cloth label \textbf{Same} k-means cluster of Color token}
  \label{fig:DCSK}
\end{subfigure}
\vspace{-5pt}
\caption{\textbf{Clustering of Color token}: 
Images from PRCC gallery set.
(a) Two images per cluster (same outline colors), indicating color embedding clusters based on background \& illumination in case of the same clothing.
(b) Cluster include similar-looking different clothes (5,6) (ideal case) and similar-looking individuals (7,8) possibly indicating \ie ReID information leak in color embeddings (error case).
}
\vspace{-7pt}
\end{figure*}

\begin{figure}[!t]
\centering
\begin{subfigure}{0.48\linewidth}
  \centering  \includegraphics[width=1\linewidth]{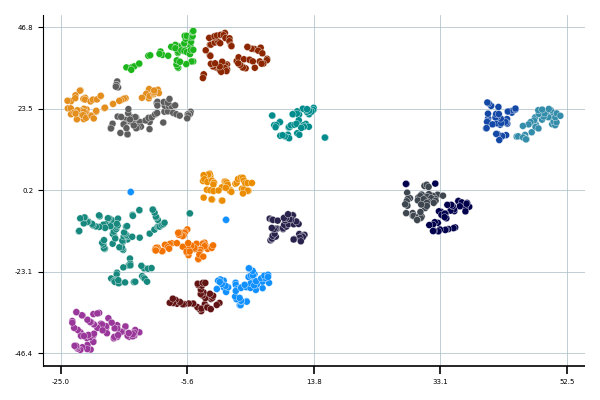}
  \caption{Clothes Labels coloring}
\label{fig:color_token_cloth_labels}
\end{subfigure}%
\hfill
\begin{subfigure}{0.52\linewidth}
  \centering  \includegraphics[width=0.92\linewidth]{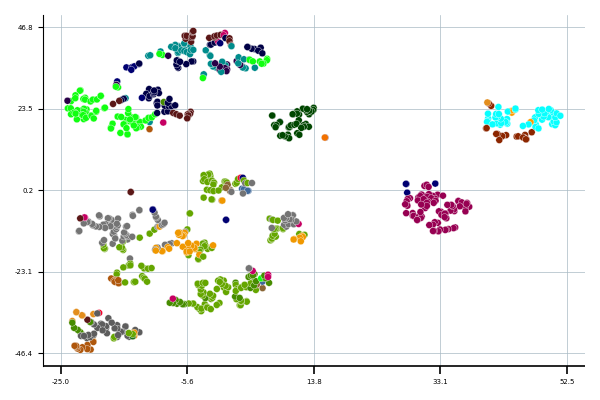}
  \caption{$f_{CO}$ K-means (k=24) coloring}
\label{fig:color_token_kmeans}
\end{subfigure}%
\caption{\textbf{Color token embeddings ($f_{CO}$) t-SNE}: 
t-SNE clusters of $f_{CO}$ can be separated into (a) clothes labels \textit{(left)} indicating a correlation between learned $f_{CO}$ and clothes labels. (b) K-means clusters of $f_{CO}$ differ in clothing \ie different clothing may share similar colors.  For example, `green' cluster on \textit{(b) right} have different clothing, blue, orange, brown on \textit{(a) left}
 (PRCC test subset)
}
\label{fig:color_token_kmeans_tsne}
\vspace{-4pt}
\end{figure}

\section{Ablation Studies \& Analysis}
All the analyses below pertain specifically to PRCC and LTCC under the CC protocol. 
Experimental details and hyperparameters of the following are in \SUPP. 

\subsection{Design Choices \& Cost Analysis}

\underline{\Cref{tab:sa_design}} demonstrates that our proposed S2A Self-Attention (\cref{fig:SA3}) outperforms the other two variants: Traditional (\cref{fig:SA1}) and Masked Self-Attention (\cref{fig:SA2}). 
The relatively lower performance of traditional self-attention on the PRCC dataset suggests an information leak between Color and ReID tokens.
CSCI parameter overhead is \textit{2.77\%} due to the Color token, its positional embeddings, and the MLP head, validating the parameter-efficient nature of using the Color token as an additional head token. 
FLOP was calculated using the fvcore library.
The increase in FLOP count for S2A Self-Attention is \textit{4.39\%} for 24 blocks, with a \textit{0.18\%} increment per block, due to the two-step self-attention process.
\underline{\Cref{tab:att_variant}} shows the additional head token-based learning for clothing labels (`w/ Clothes'), instead of color, is sub-optimal in reducing appearance bias. Grey-scaled augmentation~\cite{Li_2021_WACV} (`Grey Aug.') also performs worse, likely due to the distribution shift from RGB pre-trained weights to greyscale training. 
Using random initialization to avoid RGB bias would force training transformers on small
ReID datasets without large-scale pre-training (random initialized EVA achieves a 2.6\% Top-1 accuracy on LTCC (CC)). 
No pertaining for transformer is not practical given model’s size and complexity. 
\underline{\Cref{tab:color_design1}} indicates that input color representations (`Feed’), instead of predicting them, result in inferior performance while also introducing a test-time dependency on colors. 
Additionally, MSE regression loss ($\mathcal{L}_{MSE}^{Color}$) outperforms cosine similarity for predicting color histogram vectors.

Learning color representations instead of directly using raw color vectors offers two key benefits: 
1) Learned color representations are in the same feature space as ReID embeddings (shared transformer) which eases disentanglement. 
2) The ReID token is 1024 dimensional, while the optimal color representation of the color vector (e.g. LTCC is 3072 dim) need not fit this 1024 dimensionality. 
As shown in \underline{\cref{tab:color_token_need}} 
using sub-par color vectors (1024-dim.) or projecting ReID features to match the optimal color vectors (3072-dim.) performs inferior to learning color embeddings.

\begin{figure*}[!ht]
\centering
\begin{subfigure}{.32\linewidth}
  \centering
\includegraphics[width=0.92\linewidth]{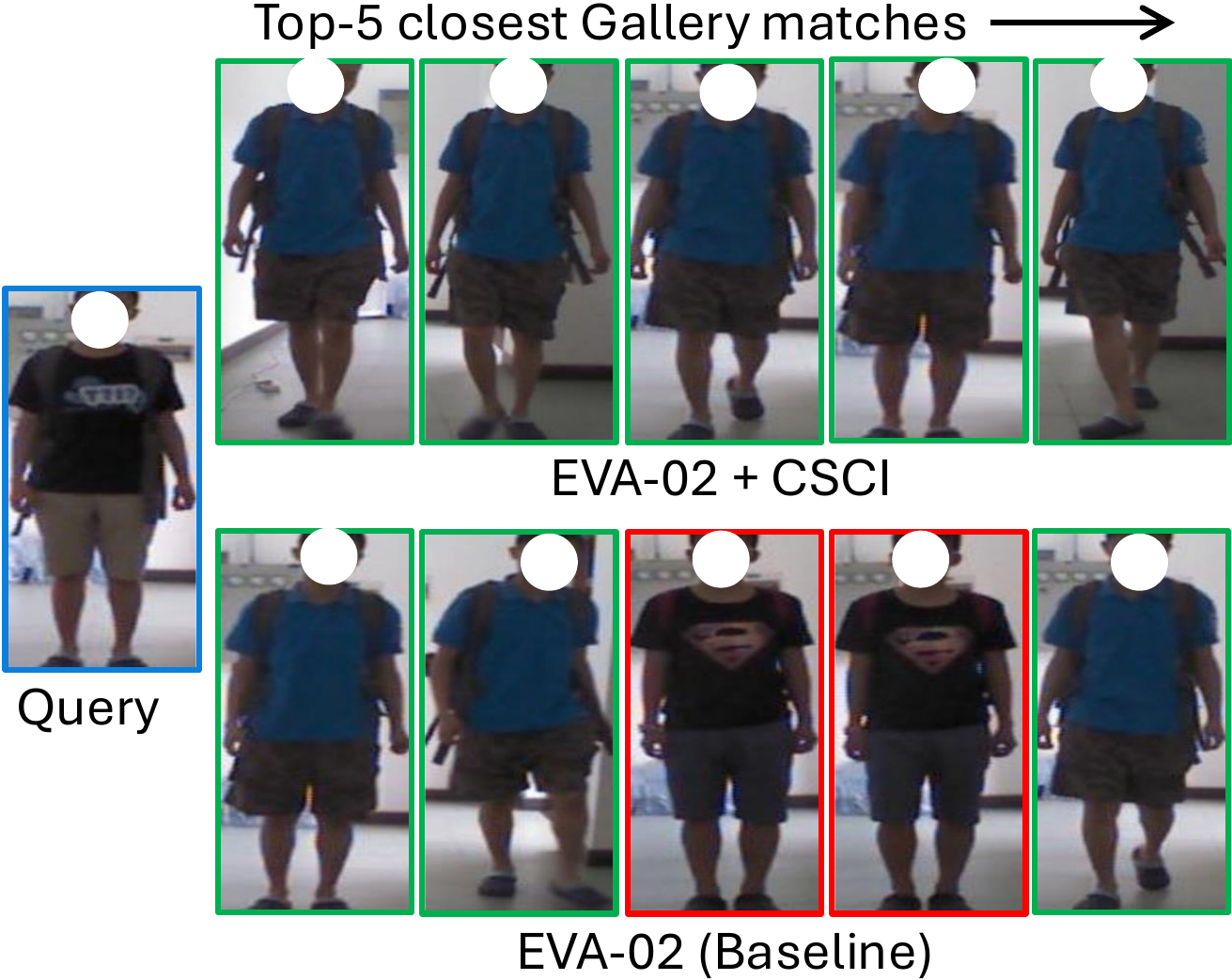}
\end{subfigure}%
\hfill
\begin{subfigure}{.32\linewidth}
  \centering
\includegraphics[width=0.92\linewidth]{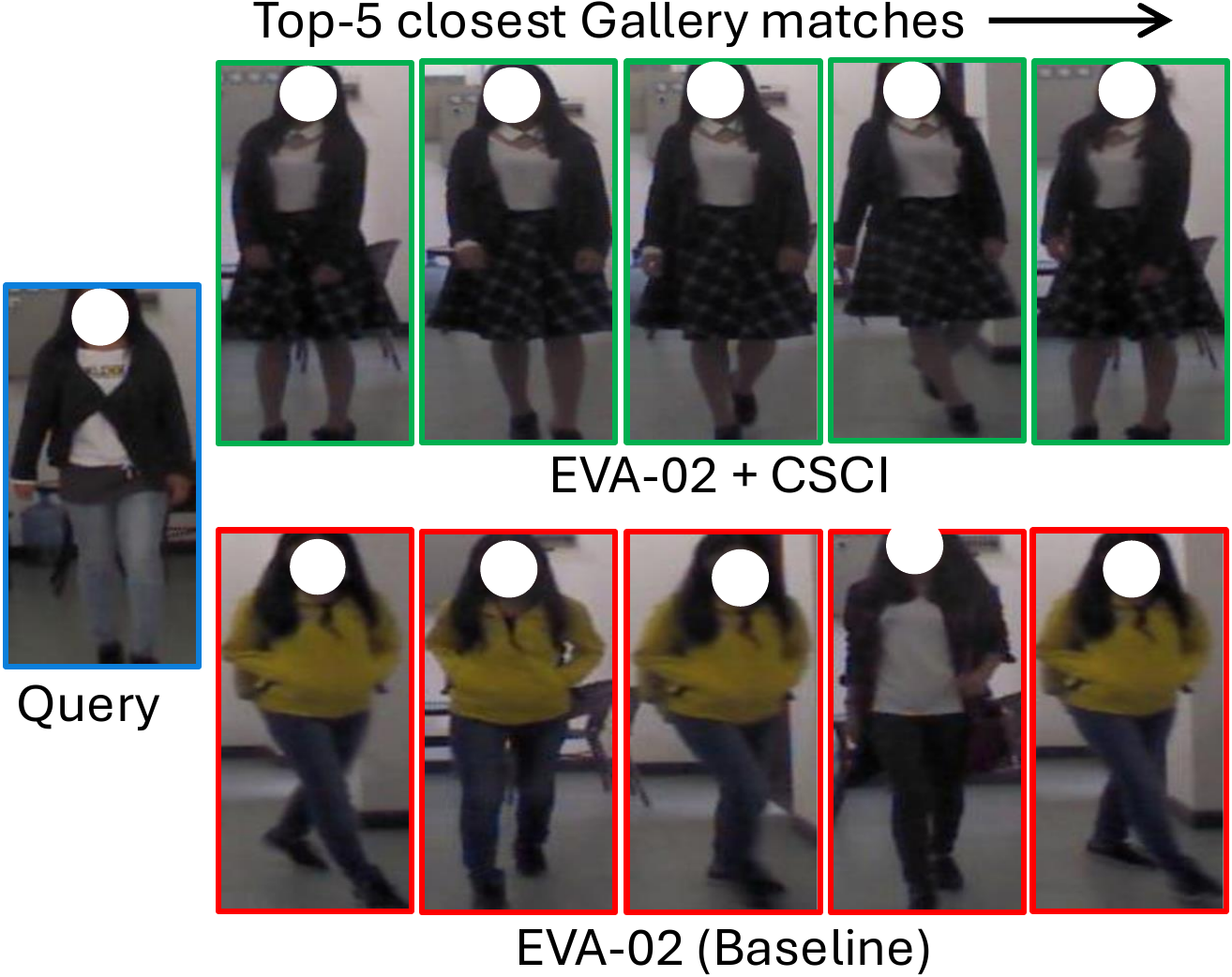}
\end{subfigure}
\hfill
\begin{subfigure}{.32\linewidth}
  \centering
\includegraphics[width=0.92\linewidth]{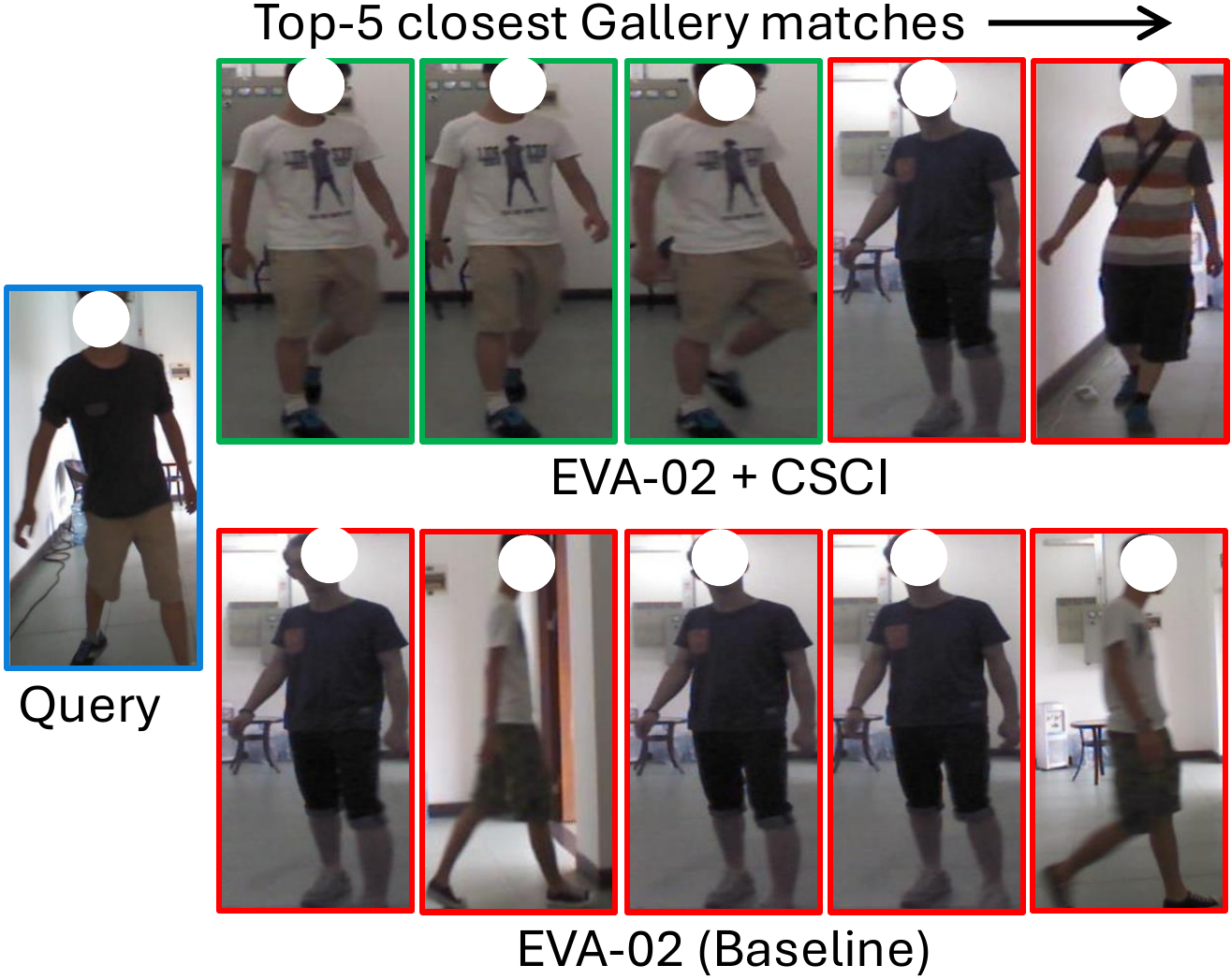}
\end{subfigure}
\vspace{-4pt}
\caption{\textbf{CSCI vs Baseline}: Column of images are top 5 gallery matches for the \BLUE{query}, taken from PRCC dataset \textbf{(clothes changing)}, with \GREEN{green} (\RED{red}) indicating \GREEN{correct} (\RED{wrong}) matches.
\textit{Left} \& \textit{Middle} are cases where baseline mistakes when the gallery wears similar clothing (\textit{left} has black t-shirt, and \textit{middle} has blue jeans). \textit{Right} shows both models making mistakes, likely because of pose similarity (orientation/tilt of hand/body). The baseline matches two different poses likely because of similar background \& illumination to the query.
}
\label{fig:match_compare}
\vspace{-3pt}
\end{figure*}

\begin{figure}[!t]
\centering
\begin{subfigure}{0.48\textwidth}
  \centering \includegraphics[width=0.85\linewidth]{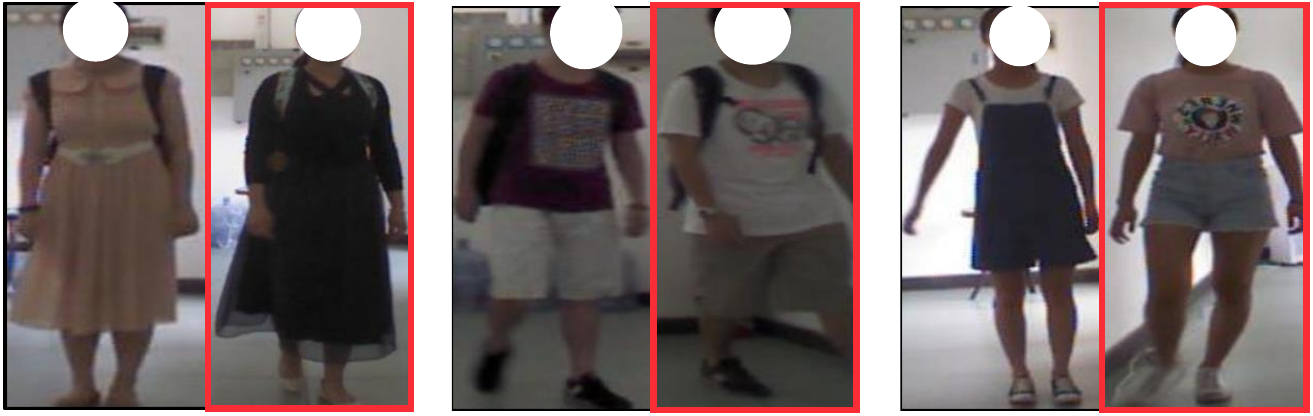}
  \caption{\textbf{Utility / Clothing Error} (\eg shorts, bag, glasses \etc)}
  \label{fig:utility_err}
\end{subfigure}
\quad 
\begin{subfigure}{0.48\textwidth}
  \centering  \includegraphics[width=0.85\linewidth]{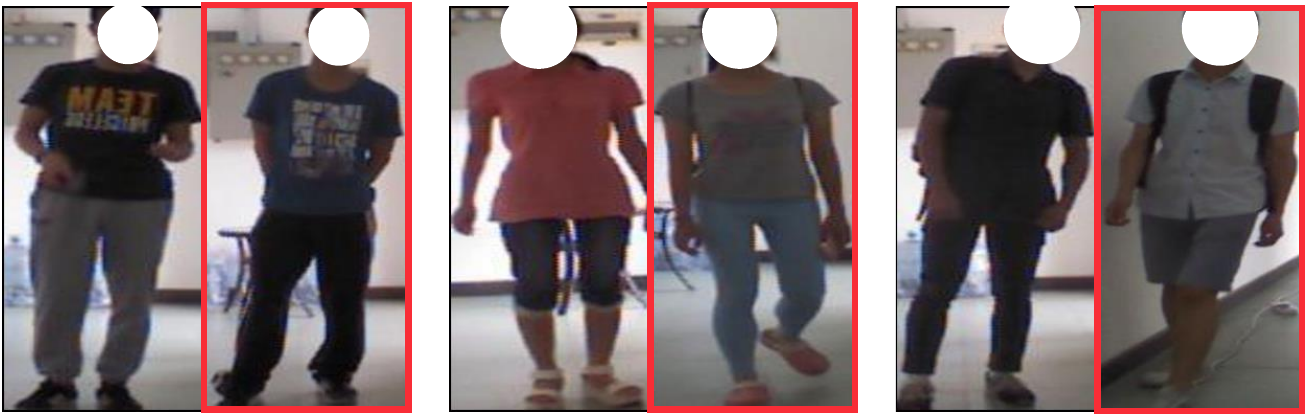}
  \caption{\textbf{Pose Error} (\eg hand position, body tile \etc)}
  \label{fig:pose_err}
\end{subfigure}
\quad 
\begin{subfigure}{0.48\textwidth}
  \centering  \includegraphics[width=0.85\linewidth]{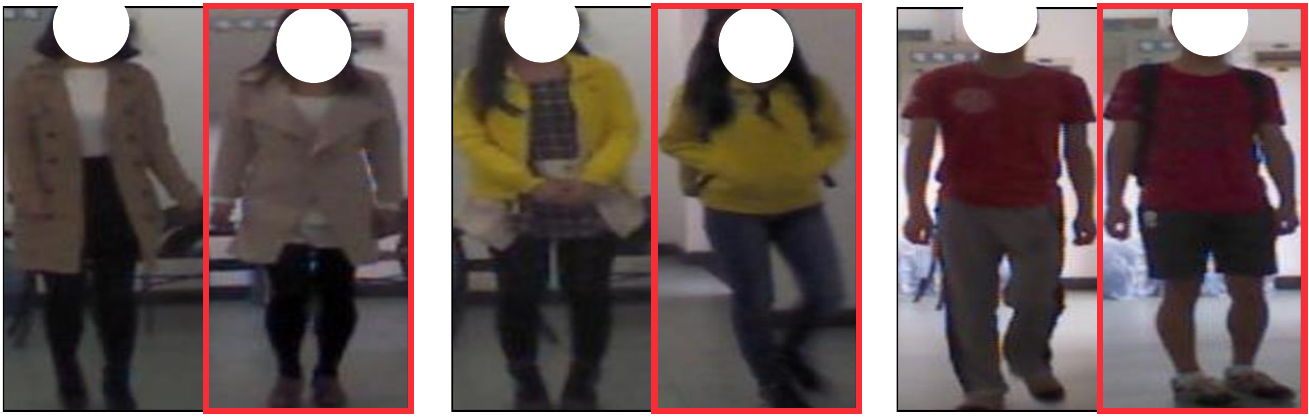}
  \caption{\textbf{Color Error} (\eg grey coat, yellow jacket, red top \etc)}
  \label{fig:color_err}
\end{subfigure}
\vspace{-16pt}
\caption{\textbf{CSCI Error analysis:} Top-1 mismatches on PRCC.
}
\label{fig:error_analysis}
\vspace{-5pt}
\end{figure}

\subsection{Color Token Embeddings}
Use of colors is motivated by its ability to distinguish between similar clothing and backgrounds. 
This is shown in \underline{\cref{fig:raw_colors_clusters}}, where RGB-uv projections create separable clusters (averaged across 3 channels, L2 normalized, t-SNE dimensionality reduction $\in\!\R^{32\x32}\!\rightarrow\!\R^{2}$, and DBSCAN). These clusters show (a) color representation can change within the same video (adaptive), (b \& c) separability based on clothing, background \& illumination, and (d) matching overall color profiles.
\textit{Supplementary} has GIFs of all clusters.

Color token embedding ($f_{CO}$) t-SNE plot using the PRCC test subset \underline{(\cref{fig:color_token_kmeans_tsne})} reveals separable clusters. 
These clusters can easily be segregated based on clothing labels \underline{(\cref{fig:color_token_cloth_labels})} 
\ie there exists a \textbf{correlation between clothing labels and learned color representation}.
Coloring these clusters via $f_{CO}$ k-means clusters \underline{(\cref{fig:color_token_kmeans})}, reveals these clusters (clothes-segregated) share color profiles across them. Simply put, different clothing labels (clusters) share learned color representation across them.

For the PRCC gallery with 71 clothing labels, k-means clustering (k=71+10 clusters) of $f_{CO}$ reveals two key findings: 
(a) \textbf{Same clothing labels but different k-means clusters:} \underline{(\cref{fig:SCDK})}, 
Similar clothes are clustered separately, likely due to variations in background and illumination.
(b) \textbf{Different clothing labels but same k-means clusters:} \underline{(\cref{fig:DCSK})}, 
Despite different clothing annotations, visually similar clothes are clustered together (5,6), indicating that $f_{CO}$ effectively identifies similar colors.
However, noisy clusters (7,8) hint at a potential information leak between $f_{ReID}$ features and $f_{CO}$ as similar poses (7) and faces (8) (hidden for privacy) are grouped together.
This issue might be solved by completely separating color embedding from ReID features (separate branch), which involves a computational efficiency trade-off (\textit{left for future research}).

\subsection{CSCI comparison with Vanilla Image Model}
The Top-5 matches for the PRCC test set were computed using our method, CSCI, and the vanilla image baseline, EVA-02, as shown in \underline{\cref{fig:match_compare}}. 
For each query (blue), we display cases where our method correctly matches (green) versus incorrect matches (red) by the baseline.
In \underline{\cref{fig:match_compare} \textit{(left \& middle)}}, the baseline makes mistakes when the gallery images share similar clothing colors with the query, such as a black t-shirt on the \textit{left} and blue jeans in the \textit{middle}. Our method, CSCI, avoids this pitfall by correctly matching distinctly different colored clothing. In \underline{\cref{fig:match_compare} \textit{(right)}}, both our model and the baseline make mistakes, likely because of similar poses such as the positioning/orientation of the hand. 
However, the baseline has two cases (2\textsuperscript{nd} \& 5\textsuperscript{th}) where the mismatch is likely due to similarity in background and illumination (rather than pose).

\subsection{Error Analysis}
\underline{\Cref{fig:error_analysis}} categorizes the Top-1 errors made by the CSCI model on the PRCC dataset into 3 types:
(a) \textbf{Utility / Clothing Errors}: Model incorrectly matches two different-looking individuals who share unique accessories like glasses (hidden with the face for privacy), a bag, or rarely appearing clothing like shorts or a one-piece skirt in the dataset. 
(b) \textbf{Similar Pose}: When query and gallery show individuals in the same pose (\eg, hand position, body tilt \etc), they are mistakenly matched, indicating a need for pose robustness. (c)\textbf{ Color Error}: Rarely appearing colors, \eg yellow or red, sometimes lead to incorrect matches, highlighting the need for improved robustness against a wide spectrum of color (clothing) appearances.



\section{Conclusion} 
\textbf{``Colors See Colors Ignore (CSCI)"} offers an RGB-only solution, based on colors as a proxy for appearance bias in real-world CC-ReID without clothing labels. 
CSCI is simple and efficient, avoiding the complexity of external models, attributes, and annotations by learning to ignore colors through the learned color embeddings. 
CSCI is shown to be superior to existing methods on both Image and Video ReID. 
We show that learned color embeddings correlate with ground-truth clothing labels and their separability under different environmental scenarios. 
CSCI hopes to offer a viable, cheaper proxy to clothing attributes/annotations.
Limitation \& ethical concerns in \SUPP.

\section{Acknowledgement}
The authors gratefully acknowledge Steven Dick (UCF High-Performance Computing) for his invaluable assistance in resolving compute-related queries, even during late-night hours.
This material is based upon work supported by the National
Science Foundation (NSF) Accelerating Research Translation (ART) program under \textit{Grant No. 2331319}.
Any opinions, findings, and conclusions or recommendations expressed in this material are those of the author(s) and
do not necessarily reflect the views of the NSF.

{
    \small
    \bibliographystyle{ieeenat_fullname}
    \bibliography{main}

\begin{thebibliography}{61}
\providecommand{\natexlab}[1]{#1}
\providecommand{\url}[1]{\texttt{#1}}
\expandafter\ifx\csname urlstyle\endcsname\relax
  \providecommand{\doi}[1]{doi: #1}\else
  \providecommand{\doi}{doi: \begingroup \urlstyle{rm}\Url}\fi

\bibitem[10.(2024)]{10.1145/3643490}
\emph{MORE '24: Proceedings of the 1st ICMR Workshop on Multimedia Object Re-Identification}, New York, NY, USA, 2024. Association for Computing Machinery.

\bibitem[Afifi and Brown(2019)]{afifi2019SIIE}
Mahmoud Afifi and Michael~S Brown.
\newblock Sensor-independent illumination estimation for dnn models.
\newblock In \emph{British Machine Vision Conference (BMVC)}, 2019.

\bibitem[Afifi et~al.(2021)Afifi, Brubaker, and Brown]{afifi2021histogan}
Mahmoud Afifi, Marcus~A. Brubaker, and Michael~S. Brown.
\newblock Histo{GAN}: Controlling colors of {GAN}-generated and real images via color histograms.
\newblock In \emph{CVPR}, 2021.

\bibitem[Ahmad et~al.(2023)Ahmad, Chanda, and Rawat]{ahmad2023ez}
Shahzad Ahmad, Sukalpa Chanda, and Yogesh~S Rawat.
\newblock Ez-clip: Efficient zeroshot video action recognition.
\newblock \emph{arXiv preprint arXiv:2312.08010}, 2023.

\bibitem[Ahmad et~al.(2025)Ahmad, Chanda, and Rawat]{ahmad2025tl}
Shahzad Ahmad, Sukalpa Chanda, and Yogesh~S Rawat.
\newblock T2l: Efficient zero-shot action recognition with temporal token learning.
\newblock \emph{Transactions on Machine Learning Research}, 2025.

\bibitem[Arkushin et~al.(2024)Arkushin, Cohen, Peleg, and Fried]{arkushin2024geff}
Daniel Arkushin, Bar Cohen, Shmuel Peleg, and Ohad Fried.
\newblock Geff: improving any clothes-changing person reid model using gallery enrichment with face features.
\newblock In \emph{Proceedings of the IEEE/CVF Winter Conference on Applications of Computer Vision}, pages 152--162, 2024.

\bibitem[Azad and Rawat(2024)]{Azad_2024_CVPR}
Shehreen Azad and Yogesh~Singh Rawat.
\newblock Activity-biometrics: Person identification from daily activities.
\newblock In \emph{Proceedings of the IEEE/CVF Conference on Computer Vision and Pattern Recognition (CVPR)}, pages 287--296, 2024.

\bibitem[Azad and Rawat(2025)]{Azad_2025_ICCV}
Shehreen Azad and Yogesh~S Rawat.
\newblock Disenq: Disentangling q-former for activity-biometrics.
\newblock In \emph{Proceedings of the IEEE/CVF International Conference on Computer Vision (ICCV)}, 2025.

\bibitem[Bansal et~al.(2022)Bansal, Foresti, and Martinel]{Bansal_2022_WACV}
Vaibhav Bansal, Gian~Luca Foresti, and Niki Martinel.
\newblock Cloth-changing person re-identification with self-attention.
\newblock In \emph{Proceedings of the IEEE/CVF Winter Conference on Applications of Computer Vision (WACV) Workshops}, pages 602--610, 2022.

\bibitem[Chen et~al.(2021)Chen, Jiang, Wang, Zhang, Zheng, Sun, and Zheng]{Chen_2021_CVPR}
Jiaxing Chen, Xinyang Jiang, Fudong Wang, Jun Zhang, Feng Zheng, Xing Sun, and Wei-Shi Zheng.
\newblock Learning 3d shape feature for texture-insensitive person re-identification.
\newblock In \emph{Proceedings of the IEEE/CVF Conference on Computer Vision and Pattern Recognition (CVPR)}, pages 8146--8155, 2021.

\bibitem[Chen et~al.(2023)Chen, Xu, Jia, Luo, Wang, Wang, Jin, and Sun]{chen2023beyond}
Weihua Chen, Xianzhe Xu, Jian Jia, Hao Luo, Yaohua Wang, Fan Wang, Rong Jin, and Xiuyu Sun.
\newblock Beyond appearance: a semantic controllable self-supervised learning framework for human-centric visual tasks.
\newblock In \emph{The IEEE/CVF Conference on Computer Vision and Pattern Recognition}, 2023.

\bibitem[Chen et~al.(2022)Chen, Xia, Zhao, Zhou, Niu, Yao, Zhu, and Liu]{CHEN202290}
Ying Chen, Shixiong Xia, Jiaqi Zhao, Yong Zhou, Qiang Niu, Rui Yao, Dongjun Zhu, and Dongjingdian Liu.
\newblock Rest-reid: Transformer block-based residual learning for person re-identification.
\newblock \emph{Pattern Recognition Letters}, 157:\penalty0 90--96, 2022.

\bibitem[Cornett et~al.(2023)Cornett, Brogan, Barber, Aykac, Baird, Burchfield, Dukes, Duncan, Ferrell, Goddard, et~al.]{cornett2023expanding}
David Cornett, Joel Brogan, Nell Barber, Deniz Aykac, Seth Baird, Nicholas Burchfield, Carl Dukes, Andrew Duncan, Regina Ferrell, Jim Goddard, et~al.
\newblock Expanding accurate person recognition to new altitudes and ranges: The briar dataset.
\newblock In \emph{Proceedings of the IEEE/CVF Winter Conference on Applications of Computer Vision}, pages 593--602, 2023.

\bibitem[Cui et~al.(2023)Cui, Zhou, Peng, Zhang, and Wang]{10036012}
Zhenyu Cui, Jiahuan Zhou, Yuxin Peng, Shiliang Zhang, and Yaowei Wang.
\newblock Dcr-reid: Deep component reconstruction for cloth-changing person re-identification.
\newblock \emph{IEEE Transactions on Circuits and Systems for Video Technology}, 33\penalty0 (8):\penalty0 4415--4428, 2023.

\bibitem[Davila et~al.(2023)Davila, Du, Lewis, Funk, Pelt, Collins, Corona, Brown, McCloskey, Hoogs, and Clipp]{Davila2023mevid}
Daniel Davila, Dawei Du, Bryon Lewis, Christopher Funk, Joseph~Van Pelt, Roderic Collins, Kellie Corona, Matt Brown, Scott McCloskey, Anthony Hoogs, and Brian Clipp.
\newblock Mevid: Multi-view extended videos with identities for video person re-identification.
\newblock In \emph{IEEE/CVF Winter Conference on Applications of Computer Vision}, 2023.

\bibitem[Devlin et~al.(2019)Devlin, Chang, Lee, and Toutanova]{Devlin2019BERTPO}
Jacob Devlin, Ming-Wei Chang, Kenton Lee, and Kristina Toutanova.
\newblock Bert: Pre-training of deep bidirectional transformers for language understanding.
\newblock In \emph{North American Chapter of the Association for Computational Linguistics}, 2019.

\bibitem[Dosovitskiy et~al.(2020)Dosovitskiy, Beyer, Kolesnikov, Weissenborn, Zhai, Unterthiner, Dehghani, Minderer, Heigold, Gelly, et~al.]{dosovitskiy2020image}
Alexey Dosovitskiy, Lucas Beyer, Alexander Kolesnikov, Dirk Weissenborn, Xiaohua Zhai, Thomas Unterthiner, Mostafa Dehghani, Matthias Minderer, Georg Heigold, Sylvain Gelly, et~al.
\newblock An image is worth 16x16 words: Transformers for image recognition at scale.
\newblock \emph{arXiv preprint arXiv:2010.11929}, 2020.

\bibitem[Fan et~al.(2023)Fan, Liang, Shen, Hou, Huang, and Yu]{Fan_2023_CVPR}
Chao Fan, Junhao Liang, Chuanfu Shen, Saihui Hou, Yongzhen Huang, and Shiqi Yu.
\newblock Opengait: Revisiting gait recognition towards better practicality.
\newblock In \emph{Proceedings of the IEEE/CVF Conference on Computer Vision and Pattern Recognition (CVPR)}, pages 9707--9716, 2023.

\bibitem[Fang et~al.(2024)Fang, Sun, Wang, Huang, Wang, and Cao]{fang2024eva}
Yuxin Fang, Quan Sun, Xinggang Wang, Tiejun Huang, Xinlong Wang, and Yue Cao.
\newblock Eva-02: A visual representation for neon genesis.
\newblock \emph{Image and Vision Computing}, page 105171, 2024.

\bibitem[Gu et~al.(2022)Gu, Chang, Ma, Bai, Shan, and Chen]{gu2022clothes}
Xinqian Gu, Hong Chang, Bingpeng Ma, Shutao Bai, Shiguang Shan, and Xilin Chen.
\newblock Clothes-changing person re-identification with rgb modality only.
\newblock In \emph{Proceedings of the IEEE/CVF conference on computer vision and pattern recognition}, pages 1060--1069, 2022.

\bibitem[Gupta and Chellappa(2024)]{Gupta_2024_WACV}
Ayush Gupta and Rama Chellappa.
\newblock You can run but not hide: Improving gait recognition with intrinsic occlusion type awareness.
\newblock In \emph{Proceedings of the IEEE/CVF Winter Conference on Applications of Computer Vision (WACV)}, pages 5893--5902, 2024.

\bibitem[Han et~al.(2023)Han, Gong, Huang, Wang, and Tan]{han2023clothing}
Ke Han, Shaogang Gong, Yan Huang, Liang Wang, and Tieniu Tan.
\newblock Clothing-change feature augmentation for person re-identification.
\newblock In \emph{Proceedings of the IEEE/CVF Conference on Computer Vision and Pattern Recognition}, pages 22066--22075, 2023.

\bibitem[He et~al.(2021)He, Luo, Wang, Wang, Li, and Jiang]{he2021transreid}
Shuting He, Hao Luo, Pichao Wang, Fan Wang, Hao Li, and Wei Jiang.
\newblock Transreid: Transformer-based object re-identification.
\newblock In \emph{Proceedings of the IEEE/CVF international conference on computer vision}, pages 15013--15022, 2021.

\bibitem[He et~al.(2024)He, Deng, Tang, Chen, Xie, Wang, Bai, Zhu, Zhao, Ouyang, Qi, and Yan]{He_2024_CVPR}
Weizhen He, Yiheng Deng, Shixiang Tang, Qihao Chen, Qingsong Xie, Yizhou Wang, Lei Bai, Feng Zhu, Rui Zhao, Wanli Ouyang, Donglian Qi, and Yunfeng Yan.
\newblock Instruct-reid: A multi-purpose person re-identification task with instructions.
\newblock In \emph{Proceedings of the IEEE/CVF Conference on Computer Vision and Pattern Recognition (CVPR)}, pages 17521--17531, 2024.

\bibitem[Jia et~al.(2022)Jia, Cheng, Lu, and Zhang]{jia2022learning}
Mengxi Jia, Xinhua Cheng, Shijian Lu, and Jian Zhang.
\newblock Learning disentangled representation implicitly via transformer for occluded person re-identification.
\newblock \emph{IEEE Transactions on Multimedia}, 25:\penalty0 1294--1305, 2022.

\bibitem[Jin et~al.(2022)Jin, He, Zheng, Yin, Shen, Huang, Feng, Huang, Chen, and Hua]{Jin_2022_CVPR}
Xin Jin, Tianyu He, Kecheng Zheng, Zhiheng Yin, Xu Shen, Zhen Huang, Ruoyu Feng, Jianqiang Huang, Zhibo Chen, and Xian-Sheng Hua.
\newblock Cloth-changing person re-identification from a single image with gait prediction and regularization.
\newblock In \emph{Proceedings of the IEEE/CVF Conference on Computer Vision and Pattern Recognition (CVPR)}, pages 14278--14287, 2022.

\bibitem[Li et~al.(2023{\natexlab{a}})Li, Wang, and Gong]{Li_2023_WACV}
Jiachen Li, Menglin Wang, and Xiaojin Gong.
\newblock Transformer based multi-grained features for unsupervised person re-identification.
\newblock In \emph{Proceedings of the IEEE/CVF Winter Conference on Applications of Computer Vision (WACV) Workshops}, pages 42--50, 2023{\natexlab{a}}.

\bibitem[Li et~al.(2024{\natexlab{a}})Li, Wang, Yu, Yan, Jia, Ding, Sheng, Liu, and Yang]{li2024rethinking}
Junjie Li, Guanshuo Wang, Fufu Yu, Yichao Yan, Qiong Jia, Shouhong Ding, Xingdong Sheng, Yunhui Liu, and Xiaokang Yang.
\newblock Rethinking clothes changing person reid: Conflicts, synthesis, and optimization.
\newblock \emph{arXiv preprint arXiv:2404.12611}, 2024{\natexlab{a}}.

\bibitem[Li et~al.(2023{\natexlab{b}})Li, Zou, Wang, Xu, Zhao, Zheng, Cheng, and Chu]{li2023dc}
Wen Li, Cheng Zou, Meng Wang, Furong Xu, Jianan Zhao, Ruobing Zheng, Yuan Cheng, and Wei Chu.
\newblock Dc-former: Diverse and compact transformer for person re-identification.
\newblock In \emph{Proceedings of the AAAI Conference on Artificial Intelligence}, pages 1415--1423, 2023{\natexlab{b}}.

\bibitem[Li et~al.(2024{\natexlab{b}})Li, Yin, Zhou, and Li]{10.1145/3643490.3661806}
Yichen Li, Yufei Yin, Wengang Zhou, and Houqiang Li.
\newblock Refining video-based person re-identification: An integrated framework with facial and body cues.
\newblock In \emph{Proceedings of the 1st ICMR Workshop on Multimedia Object Re-Identification}, New York, NY, USA, 2024{\natexlab{b}}. Association for Computing Machinery.

\bibitem[Li et~al.(2021)Li, Weng, and Kitani]{Li_2021_WACV}
Yu-Jhe Li, Xinshuo Weng, and Kris~M. Kitani.
\newblock Learning shape representations for person re-identification under clothing change.
\newblock In \emph{Proceedings of the IEEE/CVF Winter Conference on Applications of Computer Vision (WACV)}, pages 2432--2441, 2021.

\bibitem[Liang and Rawat(2025)]{Liang_2025_CVPR}
Xin Liang and Yogesh~S Rawat.
\newblock Differ: Disentangling identity features via semantic cues for clothes-changing person re-id.
\newblock In \emph{Proceedings of the Computer Vision and Pattern Recognition Conference (CVPR)}, pages 13980--13989, 2025.

\bibitem[Liu et~al.(2023)Liu, Kim, Gu, Jain, and Liu]{Liu_2023_ICCV}
Feng Liu, Minchul Kim, ZiAng Gu, Anil Jain, and Xiaoming Liu.
\newblock Learning clothing and pose invariant 3d shape representation for long-term person re-identification.
\newblock In \emph{Proceedings of the IEEE/CVF International Conference on Computer Vision (ICCV)}, pages 19617--19626, 2023.

\bibitem[Liu et~al.(2024)Liu, Kim, Ren, and Liu]{Liu_2024_CVPR}
Feng Liu, Minchul Kim, Zhiyuan Ren, and Xiaoming Liu.
\newblock Distilling clip with dual guidance for learning discriminative human body shape representation.
\newblock In \emph{Proceedings of the IEEE/CVF Conference on Computer Vision and Pattern Recognition (CVPR)}, pages 256--266, 2024.

\bibitem[Mu et~al.(2022)Mu, Li, Li, and Yang]{bmvc_lcccpr}
Jingyi Mu, Yong Li, Jun Li, and Jian Yang.
\newblock Learning clothes-irrelevant cues for clothes-changing person re-identification.
\newblock In \emph{33rd British Machine Vision Conference 2022, {BMVC} 2022, London, UK, November 21-24, 2022}, page 337. {BMVA} Press, 2022.

\bibitem[Nguyen et~al.(2024{\natexlab{a}})Nguyen, Khaldi, Nguyen, Mantini, and Shah]{Nguyen_2024_WACV}
Vuong~D. Nguyen, Khadija Khaldi, Dung Nguyen, Pranav Mantini, and Shishir Shah.
\newblock Contrastive viewpoint-aware shape learning for long-term person re-identification.
\newblock In \emph{Proceedings of the IEEE/CVF Winter Conference on Applications of Computer Vision (WACV)}, pages 1041--1049, 2024{\natexlab{a}}.

\bibitem[Nguyen et~al.(2024{\natexlab{b}})Nguyen, Mantini, and Shah]{Nguyen_2024_CVPR}
Vuong~D. Nguyen, Pranav Mantini, and Shishir~K. Shah.
\newblock Contrastive clothing and pose generation for cloth-changing person re-identification.
\newblock In \emph{Proceedings of the IEEE/CVF Conference on Computer Vision and Pattern Recognition (CVPR) Workshops}, pages 7541--7549, 2024{\natexlab{b}}.

\bibitem[Nguyen et~al.(2024{\natexlab{c}})Nguyen, Mantini, and Shah]{nguyen2024temporal}
Vuong~D Nguyen, Pranav Mantini, and Shishir~K Shah.
\newblock Temporal 3d shape modeling for video-based cloth-changing person re-identification.
\newblock In \emph{Proceedings of the IEEE/CVF Winter Conference on Applications of Computer Vision}, pages 173--182, 2024{\natexlab{c}}.

\bibitem[Ni et~al.(2023)Ni, Li, Gao, Shen, and Song]{ni2023part}
Hao Ni, Yuke Li, Lianli Gao, Heng~Tao Shen, and Jingkuan Song.
\newblock Part-aware transformer for generalizable person re-identification.
\newblock In \emph{Proceedings of the IEEE/CVF international conference on computer vision}, pages 11280--11289, 2023.

\bibitem[Pathak(2020)]{pathak2020fine}
Priyank Pathak.
\newblock Fine-grained re-identification.
\newblock \emph{arXiv preprint arXiv:2011.13475}, 2020.

\bibitem[Pathak and Rawat(2025)]{pathak2025coarse}
Priyank Pathak and Yogesh~S Rawat.
\newblock Coarse attribute prediction with task agnostic distillation for real world clothes changing reid.
\newblock \emph{arXiv preprint arXiv:2505.12580}, 2025.

\bibitem[Pathak et~al.(2020)Pathak, Eshratifar, and Gormish]{pathak2020video}
Priyank Pathak, Amir~Erfan Eshratifar, and Michael Gormish.
\newblock Video person re-id: Fantastic techniques and where to find them (student abstract).
\newblock In \emph{Proceedings of the AAAI Conference on Artificial Intelligence}, pages 13893--13894, 2020.

\bibitem[Qian et~al.(2020)Qian, Wang, Zhang, Zhu, Fu, Xiang, Jiang, and Xue]{qian2020long}
Xuelin Qian, Wenxuan Wang, Li Zhang, Fangrui Zhu, Yanwei Fu, Tao Xiang, Yu-Gang Jiang, and Xiangyang Xue.
\newblock Long-term cloth-changing person re-identification.
\newblock \emph{arXiv preprint arXiv:2005.12633}, 2020.

\bibitem[Radford et~al.(2021)Radford, Kim, Hallacy, Ramesh, Goh, Agarwal, Sastry, Askell, Mishkin, Clark, et~al.]{radford2021learning}
Alec Radford, Jong~Wook Kim, Chris Hallacy, Aditya Ramesh, Gabriel Goh, Sandhini Agarwal, Girish Sastry, Amanda Askell, Pamela Mishkin, Jack Clark, et~al.
\newblock Learning transferable visual models from natural language supervision.
\newblock In \emph{International conference on machine learning}, pages 8748--8763. PMLR, 2021.

\bibitem[Shazeer(2020)]{shazeer2020glu}
Noam Shazeer.
\newblock Glu variants improve transformer.
\newblock \emph{arXiv preprint arXiv:2002.05202}, 2020.

\bibitem[Shu et~al.(2021)Shu, Li, Wang, Ruan, and Tian]{shu2021semantic}
Xiujun Shu, Ge Li, Xiao Wang, Weijian Ruan, and Qi Tian.
\newblock Semantic-guided pixel sampling for cloth-changing person re-identification.
\newblock \emph{IEEE Signal Processing Letters}, 28:\penalty0 1365--1369, 2021.

\bibitem[Su et~al.(2024)Su, Ahmed, Lu, Pan, Bo, and Liu]{su2024roformer}
Jianlin Su, Murtadha Ahmed, Yu Lu, Shengfeng Pan, Wen Bo, and Yunfeng Liu.
\newblock Roformer: Enhanced transformer with rotary position embedding.
\newblock \emph{Neurocomputing}, 568:\penalty0 127063, 2024.

\bibitem[Touvron et~al.(2021)Touvron, Cord, Douze, Massa, Sablayrolles, and J{\'e}gou]{touvron2021training}
Hugo Touvron, Matthieu Cord, Matthijs Douze, Francisco Massa, Alexandre Sablayrolles, and Herv{\'e} J{\'e}gou.
\newblock Training data-efficient image transformers \& distillation through attention.
\newblock In \emph{International conference on machine learning}, pages 10347--10357. PMLR, 2021.

\bibitem[Wan et~al.(2020)Wan, Wu, Qian, Chen, and Fu]{Wan_2020_CVPR_Workshops}
Fangbin Wan, Yang Wu, Xuelin Qian, Yixiong Chen, and Yanwei Fu.
\newblock When person re-identification meets changing clothes.
\newblock In \emph{Proceedings of the IEEE/CVF Conference on Computer Vision and Pattern Recognition (CVPR) Workshops}, 2020.

\bibitem[Wang et~al.(2022{\natexlab{a}})Wang, Ma, Huang, Dong, Wang, Peng, Wu, Bajaj, Singhal, Benhaim, et~al.]{wang2022foundation}
Hongyu Wang, Shuming Ma, Shaohan Huang, Li Dong, Wenhui Wang, Zhiliang Peng, Yu Wu, Payal Bajaj, Saksham Singhal, Alon Benhaim, et~al.
\newblock Foundation transformers.
\newblock \emph{arXiv preprint arXiv:2210.06423}, 2022{\natexlab{a}}.

\bibitem[Wang et~al.(2022{\natexlab{b}})Wang, Zhang, Han, Yang, Li, Feng, and Wang]{wang2022benchmark}
Likai Wang, Xiangqun Zhang, Ruize Han, Jialin Yang, Xiaoyu Li, Wei Feng, and Song Wang.
\newblock A benchmark of video-based clothes-changing person re-identification.
\newblock \emph{arXiv preprint arXiv:2211.11165}, 2022{\natexlab{b}}.

\bibitem[Wang et~al.(2024{\natexlab{a}})Wang, Qian, Li, Fu, and Xue]{wang2024image}
Qizao Wang, Xuelin Qian, Bin Li, Yanwei Fu, and Xiangyang Xue.
\newblock Image-text-image knowledge transferring for lifelong person re-identification with hybrid clothing states.
\newblock \emph{arXiv preprint arXiv:2405.16600}, 2024{\natexlab{a}}.

\bibitem[Wang et~al.(2024{\natexlab{b}})Wang, Qian, Li, Xue, and Fu]{wang2024exploring}
Qizao Wang, Xuelin Qian, Bin Li, Xiangyang Xue, and Yanwei Fu.
\newblock Exploring fine-grained representation and recomposition for cloth-changing person re-identification.
\newblock \emph{IEEE Transactions on Information Forensics and Security}, 2024{\natexlab{b}}.

\bibitem[Wang et~al.(2021)Wang, Zhang, Gao, Geng, Lu, and Wang]{9710753}
Yingquan Wang, Pingping Zhang, Shang Gao, Xia Geng, Hu Lu, and Dong Wang.
\newblock Pyramid spatial-temporal aggregation for video-based person re-identification.
\newblock In \emph{2021 IEEE/CVF International Conference on Computer Vision (ICCV)}, pages 12006--12015, 2021.

\bibitem[Wang et~al.(2023)Wang, Jiang, Xu, and Sun]{wang2023transformer}
Zepeng Wang, Xinghao Jiang, Ke Xu, and Tanfeng Sun.
\newblock A transformer-based cloth-irrelevant patches feature extracting method for long-term cloth-changing person re-identification.
\newblock In \emph{Advances in Computer Graphics: 39th Computer Graphics International Conference, CGI 2022, Virtual Event, September 12--16, 2022, Proceedings}, pages 278--289. Springer, 2023.

\bibitem[Wu et~al.(2020)Wu, Bourahla, Li, Wu, Tian, and Zhou]{wu2020adaptive}
Yiming Wu, Omar El~Farouk Bourahla, Xi Li, Fei Wu, Qi Tian, and Xue Zhou.
\newblock Adaptive graph representation learning for video person re-identification.
\newblock \emph{IEEE Transactions on Image Processing}, 29:\penalty0 8821--8830, 2020.

\bibitem[Xiong et~al.(2024)Xiong, Yang, Chen, Aly, AlTameem, Saudagar, Mumtaz, and Muhammad]{10497899}
Mingfu Xiong, Xinxin Yang, Hanmei Chen, Wael Hosny~Fouad Aly, Abdullah AlTameem, Abdul Khader~Jilani Saudagar, Shahid Mumtaz, and Khan Muhammad.
\newblock Cloth-changing person re-identification with invariant feature parsing for uavs applications.
\newblock \emph{IEEE Transactions on Vehicular Technology}, pages 1--10, 2024.

\bibitem[Yang et~al.(2019)Yang, Wu, and Zheng]{yang2019person}
Qize Yang, Ancong Wu, and Wei-Shi Zheng.
\newblock Person re-identification by contour sketch under moderate clothing change.
\newblock \emph{IEEE transactions on pattern analysis and machine intelligence}, 43\penalty0 (6):\penalty0 2029--2046, 2019.

\bibitem[Yang et~al.(2023)Yang, Lin, Zhong, Wu, and Wang]{Yang_2023_CVPR}
Zhengwei Yang, Meng Lin, Xian Zhong, Yu Wu, and Zheng Wang.
\newblock Good is bad: Causality inspired cloth-debiasing for cloth-changing person re-identification.
\newblock In \emph{Proceedings of the IEEE/CVF Conference on Computer Vision and Pattern Recognition (CVPR)}, pages 1472--1481, 2023.

\bibitem[Ye et~al.(2024)Ye, Chen, Li, Zheng, Crandall, and Du]{ye2024transformer}
Mang Ye, Shuoyi Chen, Chenyue Li, Wei-Shi Zheng, David Crandall, and Bo Du.
\newblock Transformer for object re-identification: A survey.
\newblock \emph{arXiv preprint arXiv:2401.06960}, 2024.

\bibitem[Zhu et~al.(2024)Zhu, Zheng, Zheng, and Nevatia]{Zhu_2024_WACV}
Haidong Zhu, Wanrong Zheng, Zhaoheng Zheng, and Ram Nevatia.
\newblock Sharc: Shape and appearance recognition for person identification in-the-wild.
\newblock In \emph{Proceedings of the IEEE/CVF Winter Conference on Applications of Computer Vision (WACV)}, pages 6290--6300, 2024.

\end{thebibliography}
}

\maketitlesupplementary
\section*{APPENDIX}

In this supplementary we have included:\\
\noindent \textbf{1)} \textbf{Intuition behind using colors} justification of why colors can be a good lightweight substitute for clothing labels in reducing appearance bias. (\Cref{sec:color_reason}).\\
\noindent \textbf{2)} \textbf{Description of two Zips} attached with this supplementary, the RGB GIFs denoting the clusters based on color representation (\Cref{sec:rgb_gifs}).\\
\noindent \textbf{3) RGB-uv Histogram Mathematical description} (\Cref{sec:rgbuv_hist}).\\
\noindent \textbf{4) Color Hyperparamters} All color profiles related hyperparameters used in the main paper (\Cref{sec:color_hist_params}). \\
\noindent \textbf{5) Image Encoder EVA-02} All color profiles related hyperparameters used in the main paper (\Cref{sec:eva_details}).\\
\noindent \textbf{6) EZ-CLIP} Video ReID model architecture (\Cref{sec:ez_clip}).\\
\noindent \textbf{7) Intutive explanation of S2A self-attention} (\Cref{sec:s2a_intutive}).\\
\noindent \textbf{8) Gram CAM Comparison} Grad cam visualization for CSCI and comparison with baseline. (\Cref{sec:grad_cam}).\\
\noindent \textbf{9) All Losses} Formula for all the loss function used to train models. 
(\Cref{sec:loss_formulations}).\\
\noindent \textbf{10) All Results} Table include technique based previous works which can are generally applied on pre-trained models (\Cref{sec:all_results}).\\
\noindent \textbf{11) Mevid Results } All Mevid evaluation protocols, namely overall, clothing-based, distance/location-based, and scale-based.  (\Cref{sec:mevid_protocols}).\\
\noindent \textbf{12) Limitation and ethical consideration} (\Cref{sec:limitations}).\\

\section{Intuition behind using colors}
\label{sec:color_reason}

Colors’s inherently have limitations like lacking global context with noisy/ambiguity.
However, our goal is not to solve/replace clothing, but rather to offer a lightweight/practical alternative to costly clothing labels/annotation.
Notably, existing clothing labels also suffer similar challenges of not modeling clothing style, structure, or texture and disregarding spatial and structural information, since they rely on a single integer label to represent clothing (\eg ``cloth 070 id 1")~\cite{gu2022clothes, Yang_2023_CVPR, Nguyen_2024_CVPR}.

Currently, addressing these requires richer descriptions via LLMs or diffusion models (IRM~\cite{He_2024_CVPR}), computationally impractical. 
Moreover, LLMs would need to track clothing changes over time, while integer labels only assign one noisy integer label for the entire video.
In contrast, colors are more expressive than integer labels, outperforming integer labels: +2.3\% for ViT-B, and +1.5\% for EVA-02 (Tab.3,4(b)). Compared to LLMs, colors are more cost-effective.

\textbf{Colors seems like an intuitive proxy for clothing, effectively clustering images with similar clothing} (Fig. 2, \& next section).
On real-world, complex clothing dataset, MeVID, colors shows improvement of +3.6\%.

\section{RGB GIFs (Figure 1)}
\label{sec:rgb_gifs}
Figure 1 main submission shows CCVID video frames and LTCC images clustered using RGB-uv projections. 
We used the DBSCAN algorithm to cluster color histograms since we don't know the number of actual clusters. In the main paper, we have shown 7 clusters, while the number of actual clusters is 943 for CCVID. 
Supplementary has two \textit{`CCVID Color Clusters (Fig 1)'} and \textit{'LTCC Color Clusters (Fig 1)'}
has the gifs of these clusters (we have taken ~3-5 images from each cluster).

\section{RGB-uv Histograms derivation \textit{(Under 
Color Representation in Methodology)}}
\label{sec:rgbuv_hist}
HistGAN~\cite{afifi2021histogan} used RGB-uv projection of colors for style transfer in GANs.
They implied this representation (2D representation) has better invariance to illumination, is more compact than RGB pixel binning (3D), and can easily be generated via a set of differential equations, making them compatible with training models. 

This representation is a 2D histogram image projection of R,G,B colors in \textit{u-v} log-chroma space.
\textbf{These $\textit{u-v}$ 2D axis are bins of histogram denoted by $h$ in the main paper. }
The log-chroma space is
defined by the intensity of one channel ($I_R, I_G, I_B$), normalized by the
other two channels on two dimensions (`uv', $I_u$, and $I_v$), as shown below for red channel. 
\begin{align}
    I_{\underline{u}R}(x)=\log{\big(\frac{I_R(x)+\epsilon}{I_G(x)+\epsilon}\big)} \\
    I_{\underline{v}R}(x)=\log{\big(\frac{I_R(x)+\epsilon}{I_B(x)+\epsilon} \big)} 
    \label{eq:uv_projection}
\end{align}
where x is the pixel index, $I_R(x)$ is the pixel intensity in channel R and $\epsilon$ is small stability constant. Similarly green and blue channels can be projected as $[I_{uG}, I_{vG}]$, and $[I_{uB}, I_{vB}]$. 
The histogram H in 2D space ($u,v$) for channel $c$ is computed as 
\begin{align}
    \underline{I_y(x)} &= \sqrt{I^2_R(x) + I^2_G(x) + I^2_B(x)}
\end{align}
\begin{align}    
    \textbf{k}(I_{uc}(x), I_{vc}(x), u, v) &=  \Bigg( 1+ \bigg(\frac{|I_{uc}(x) - u |}{\tau} \bigg)^2 \Bigg)^{-1} \times \nonumber \\
    & \hspace{1cm} \Bigg( 1+ \bigg(\frac{|I_{vc}(x) - v |}{\tau} \bigg)^2 \Bigg)^{-1} 
\end{align}
\begin{align}    
    H(u, v, c) &\propto \sum_x \textbf{k}(I_{uc}(x), I_{vc}(x), u, v) \underline{I_y(x)} \\
    \sum_{u,v,c}  H(u, v, c) &= 1
\end{align}
where $k(I_{uc}(x), I_{vc}(x), u, v)$ is called the inverse-quadratic kernel, used for all our experiments, and $\tau$ is a fall-off parameter to control the smoothness of the histogram’s bins.

\section{Color Histograms hyperparameters \textit{(Under 
Implementation in Experiments)}
}
\label{sec:color_hist_params}
Color histogram hyperparameters differ such as reporting the best accuracy obtained for a particular experiment. 
These hyperparameters are described below with ones used in experimental values reported in the main submission listed in \Cref{tab:color_hyperparam_exps}.

\begin{table*}
\centering
\begin{tabular}{|l|P{1cm}|P{1.5cm}|c|P{1.7cm}|P{1.3cm}|P{1.7cm}|c|}
\toprule
Exp. / Fig. & Dataset &Type of Histogram & Bin Size & Smoothness $\tau$ & Concat / Mean & Normalization L2 / L1 & Scaling \\ 
\hline 
\toprule
Figure 1 & CCVID & RGB-uv & 32 & 0.02 & Mean & L2 & 1\\ 
\hline
\multirow{2}{*}{Table 2 EZ+CSCI} & \multirow{2}{*}{CCVID} & Pixel Bin & 20 & - &  & - & -  \\ 
  &  &  RGB-uv & 32 & 0.02 & - & - & 100  \\ 
\hline
\multirow{4}{*}{Table 1 CSCI}
 & \multirow{2}{*}{LTCC} & Pixel Bin & 20 & - & Concat & L2 & 100  \\ 
  & & RGB-uv & 32 & 0.001 & Concat & L2 & 100  \\ 
\cline{2-8}
  & \multirow{2}{*}{PRCC} & Pixel Bin & 20 & - & Concat & L1 & 1000  \\ 
  & & RGB-uv & 32 & 0.001 & Concat & L1 & 10  \\ 
\hline
\multirow{4}{*}{Table 2 EZ+CSCI}
& \multirow{2}{*}{CCVID} & Pixel Bin & 20 & - & Concat & L2 & 1000  \\ 
& & RGB-uv & 16 & 0.001 & Mean & min / max & 10  \\ 
\cline{2-8}
 & \multirow{2}{*}{MEVID}& Pixel Bin & 32 & - & Concat & L2 & 10  \\ 
 & & RGB-uv & 32 & - & Mean & min / max & 1  \\
\hline 
Table 3 (a) Traditional  &  \multirow{3}{*}{LTCC} & RGB-uv & 32 & 0.002 & Concat & min / max & 1000  \\ 
Table 3 (a) Masked  &  & RGB-uv & 32 & 0.002 & Mean & min / max & 10  \\ 
Table 3 (a) S2A  &  & RGB-uv & 32 & 0.001 & Concat & L2 & 100  \\ 
\hline 
Table 3 (a) Traditional  &  \multirow{3}{*}{PRCC} & RGB-uv & 16 & 0.002 & Concat & min / max & 1000  \\ 
Table 3 (a) Masked  &  & RGB-uv & 32 & 0.002 & Mean & L1 & 1  \\ 
Table 3 (a) S2A  &  & RGB-uv & 32 & 0.001 & Concat & L1 & 10  \\
\hline 
Table 4 TransReID &  \multirow{4}{*}{LTCC} & RGB-uv & 64 & 0.001 & Mean & L1 & 10 \\ 
Table 4 TMGF  &  & RGB-uv & 32 & 0.001 & Concat & L2 & 100  \\ 
Table 4 PAT   &  & RGB-uv & 32 & 0.001 & Concat & min / max & 10  \\
Table 4 TCiP   &  & RGB-uv & 32 & 0.002 & Concat & min / max & 1000  \\
\bottomrule
\end{tabular}
\caption{\textbf{Hyperparams for Color histograms}: For each experiment and figure in the main submission. `-' means a particular hyparameter that is not applicable.}
\label{tab:color_hyperparam_exps}
\end{table*}

\subsection{RGB-uv histograms}
The RGB-uv histograms $H(u, v, c)$ has 5 set of hyperparameters associated with it: 
\begin{enumerate}
\item \textbf{Bin size $h=u=v$}: size of histogram images. $h=32$ is shown for all demonstration purposes in the main submission.
\item \textbf{Smoothness factor $\tau$}: $\tau = 0.02$ is used for all demonstration purposes in the main submission. 
\item \textbf{Concatenating vs Averaging the 3 channels}. We flatten these histograms to regress over via color tokens. Hence we can either concatenate these channels by creating $R^{3\times h\times h}$ 1D vector or averaging $R^{h\times h}$ vectors.  
\item \textbf{Normalization}: Normalized flattened histogram vector is fed to the model. It can be either L2 normalized or L1 normalized or simple min, max normalization $\frac{x - min(x)}{max(x) - min(x)}$
\item \textbf{Scale Multiplication factor}. Once normalized most of the values of histograms are close to 0, \eg $3 \times 32\times32$ L2 normalized vector. Hence, we multiply the vectors by 1, 10, 100, or 1000 as a scale factor.  
\end{enumerate}

\begin{figure}[!t]
\centering
\centering 
\includegraphics[width=0.98\linewidth]{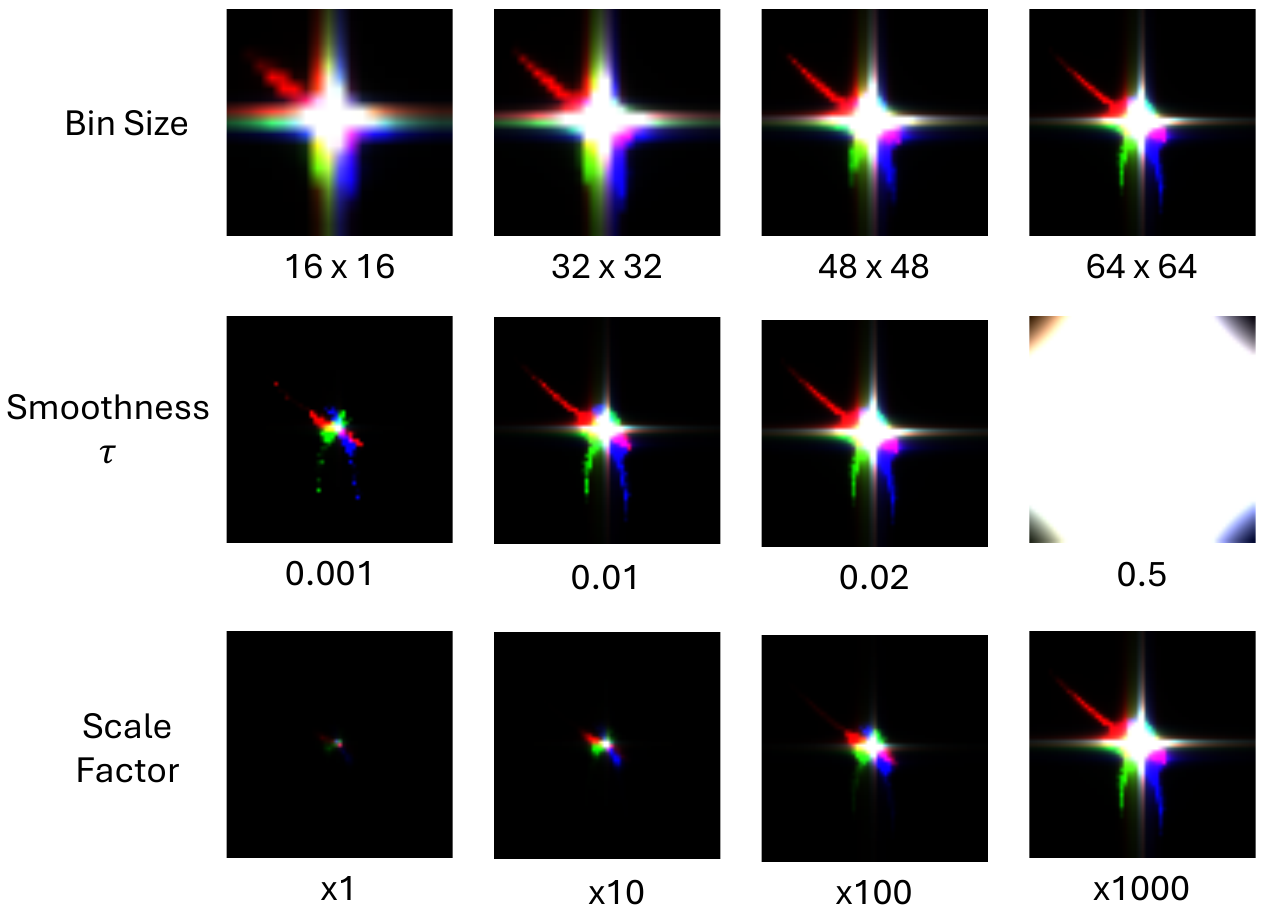}
\caption{\textbf{RGB-uv parameters:} 
All images are resized for better visualization 
(a) \textit{Bin size} is visualized using a scale factor of 1000 and smoothness of 0.02. 
(b) \textit{Smoothness Factor $\tau$} is calculated for 64 bin size and scale factor 1000.
(c) \textit{Scale factor} is calculated for 64 bin size and smoothness of 0.02.
}
\label{fig:rgb_uv_hyperparam_viz}
\end{figure}

These hyperparameters namely, bin size, smoothness factor, and scale factor are visualized in \cref{fig:rgb_uv_hyperparam_viz}.     

\subsection{Pixel Binning histograms}

Pixel binning produces a 3D $h\times h\times h$ histogram, with bin size $h$. Its always concatenated, producing $R^{h\times h\times h}$ 1D flattened vector.
It has 3 sets of hyperparameters. 
\begin{enumerate}
\item \textbf{Bin size $h$}: size of histogram bins. $h=20$ is constant for all the experiments, producing $20\times20\times20$ 3D histogram.
\item \textbf{Normalization}, same as RGB-uv histograms
\item \textbf{Scale Multiplication factor}
same as RGB-uv histograms.
\end{enumerate}

\subsection{Color Hyper parameter performance}
\begin{figure}[!t]
\centering
\centering 
\includegraphics[width=0.98\linewidth]{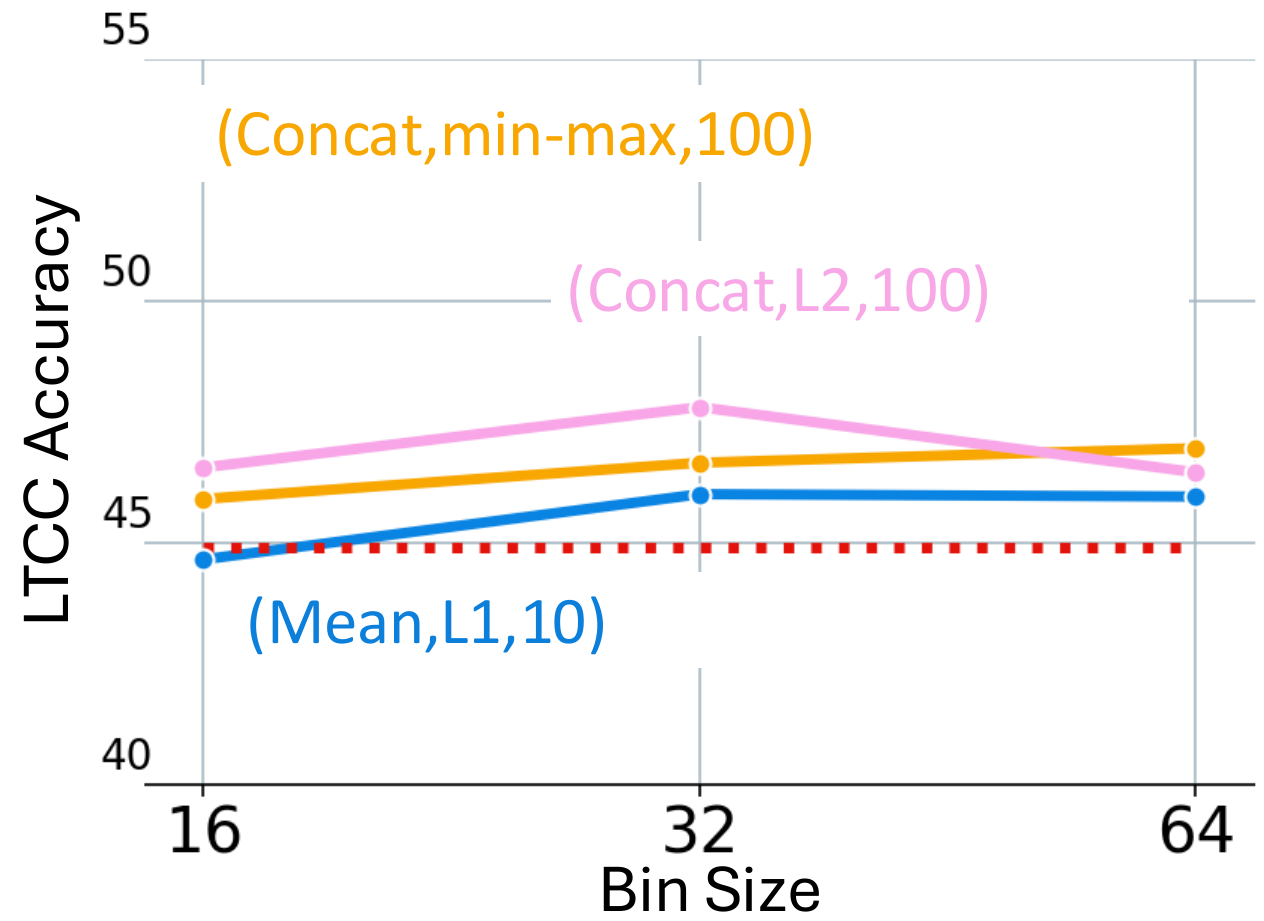}
\caption{\textbf{LTCC performance for various RGB-uv hyper parameters:} 
The red dotted line indicates the baseline @44.9\%. The color-coded text indicates the hyperparameters as (Concate/Mean, normalization, and scaling weight).  All plots have $\tau=0.001$. 
}
\label{fig:color_performance_hyperparam}
\end{figure}

As shown in \cref{tab:color_hyperparam_exps}, there are several hyperparameters. In order to give some estimate of the relative performance of each hyperparameters, we have plotted LTCC performance for some of them using RGB-uv projections with smoothing factor $\tau=0.001$ for all bin sizes, as shown in \cref{fig:color_performance_hyperparam}. 
All hyperparams have more or less the same performance, with an average of 2 runs reported. The peak performance was obtained in the order of $\sim49\%$, averaging around $47.80$ for the concatenation of three channels, with L2 normalization, a scaling factor of 100, and bin size=32.

\section{Visual Transformer Encoder EVA-02 Details 
\textit{(Under Color See Color Ignore (CSCI)
in Methodology)}
}
\label{sec:eva_details}
\begin{figure}[!t]
\centering
\centering 
\includegraphics[width=0.8\linewidth]{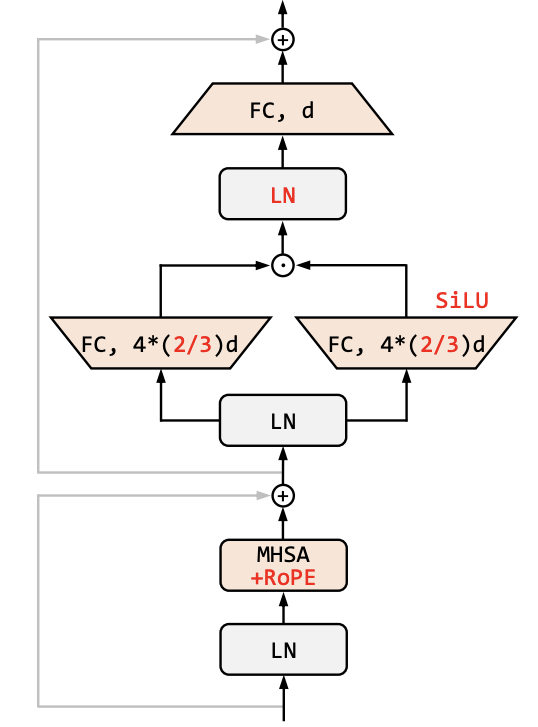}
\caption{\textbf{EVA-02 self-attention with RoPE Emebeddings} 
Figure is taken from ~\citeauthor{fang2024eva}.
}
\label{fig:eva_sa}
\end{figure}

EVA-02~\cite{fang2024eva} is a vision-language transformer with large-scale CLIP based pre-training~\cite{radford2021learning}. 
It has shown remarkable generalization with state-of-the-art zero-shot image classification capabilities.
At the backend, EVA-02 has a series of `modified' ViT-Large blocks~\cite{dosovitskiy2020image}.
These modifications include feed-forward network \textit{SwiGLU}~\cite{shazeer2020glu}, layer normalization \textit{sub-LN}~\cite{wang2022foundation}, and rotary position embedding \textit{RoPE}~\cite{su2024roformer})\footnote{EVA02 architectural details: \url{https://arxiv.org/pdf/2303.11331}}. 
The model operates like a ViT transformer, converting the image into non-overlapping spatial tokens, appended with a class token. These tokens (along with their positional embeddings) are then passed through a series of EVA-02 (ViT) blocks for multi-head self-attention. This self-attention is displayed in the \cref{fig:eva_sa} where the above-mentioned modifications are applied. Instead of traditional MLP, the SwiGLU feed-forward network is used. 
Additionally, RoPE embeddings are before the self-attention step. 
Instead of traditional layer normalization \textit{sub-LN} is applied. 
The additional fraction $\frac{2}{3}$ dimensionality keeps the number of parameters and FLOPs counts the same as ViT.

\begin{figure}[!t]
\centering
\includegraphics[width=0.92\columnwidth]{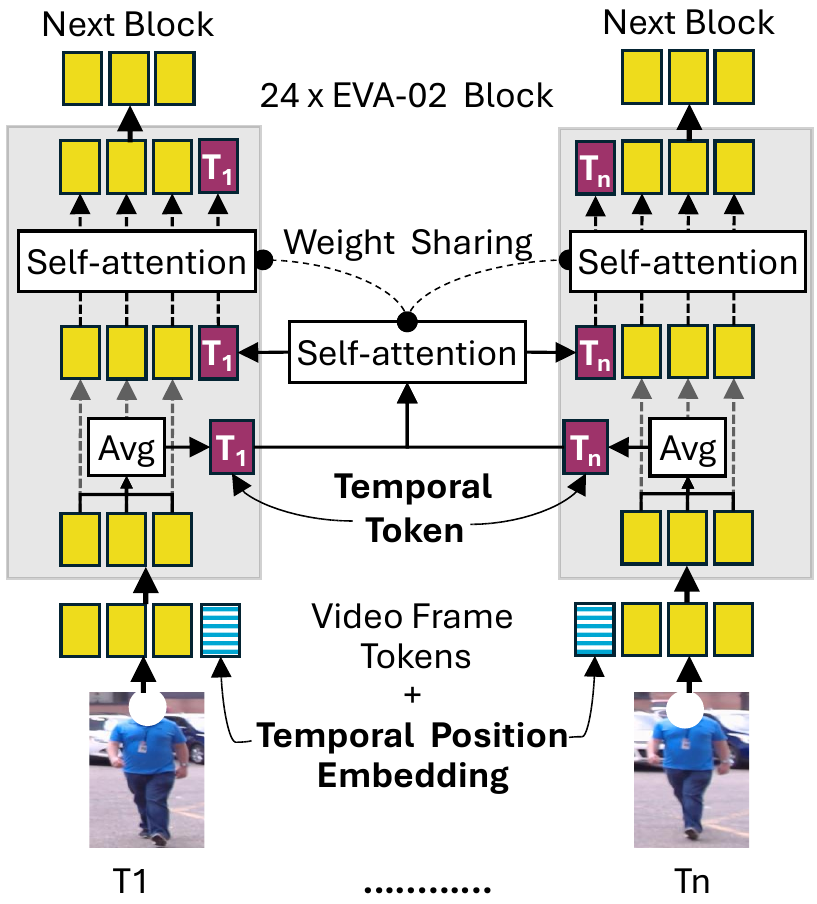} 
\caption{ \textbf{EZ-CLIP (Video ReID)}: EVA-02 Video architecture.}
\label{fig:ezclip}
\end{figure}

\section{Video ReID : EZ-CLIP : (EC)
\textit{(Under Generalization to Video ReID
in Methodology)}
}
\label{sec:ez_clip}
Recently \cite{ahmad2023ez} proposed using temporal prompts (tokens) for converting any image-based transformers into a video model\footnote{EZ-CLIP code:\url{https://github.com/Shahzadnit/EZ-CLIP}}. 
We first pre-train the EVA-02 image encoder on the video dataset,  using random frames for Image ReID. 
Then, the pre-trained image model is applied to each video frame ($n=4$ frames). 
`Temporal' position embeddings (blue) are added to the `spatial' video frame tokens (yellow) for each timestamp as shown \cref{fig:ezclip}. 
Within each EVA-02 block, the forward pass has two steps: 1) 
the mean of spatial tokens is added to a newly introduced trainable temporal token (orange) for each timestamp. 
These temporal tokens undergo self-attention using the shared weight of the self-attention layer of spatial tokens.
2) Self-attended temporal tokens are appended to the spatial tokens, similar to class tokens for standard self-attention, i.e. the number of tokens increases by one for each timestamp.
After self-attention, temporal tokens are discarded and spatial tokens are forwarded to the next block for identical two-stage forward pass. 
All model weights are shared between spatial and temporal tokens, with the temporal tokens being the only new parameters added. 
The entire network is trained end-to-end during the video phase, with temporal pooling on the class token to obtain the final $f_{ReID}$, effectively introducing temporal context into spatial tokens. 
To reduce redundancy across frames Motion loss ($\mathcal{L}_{ML}$) is introduced as,
\begin{align}
    C_{mean} &= \frac{1}{n} \sum_i^{n} \big(f^i_{ReID} - f^{i+1}_{ReID}\big)\\
    Var &= \frac{1}{n} \sum_i^{n} \big(f^i_{ReID} - \frac{1}{n} \sum_i^{n} f^i_{ReID}\big)^2\\
    \mathcal{L}_{ML} &= \frac{1}{C+V}
    \label{eq:ml_loss}
\end{align}
Here $\mathcal{L}_{ML}$ penalizes small $C_{mean}$ (difference between consecutive frames) and $Var$ (variance across the frames).

Using the pre-trained Image Encoder, one epoch employs temporal tokens for Video ReID, and in the \textit{alternate} epochs, color tokens for disentangling colors as Image ReID, (middle frame of video as input).

\begin{figure}[!t]
\centering
\includegraphics[width=0.98\columnwidth]{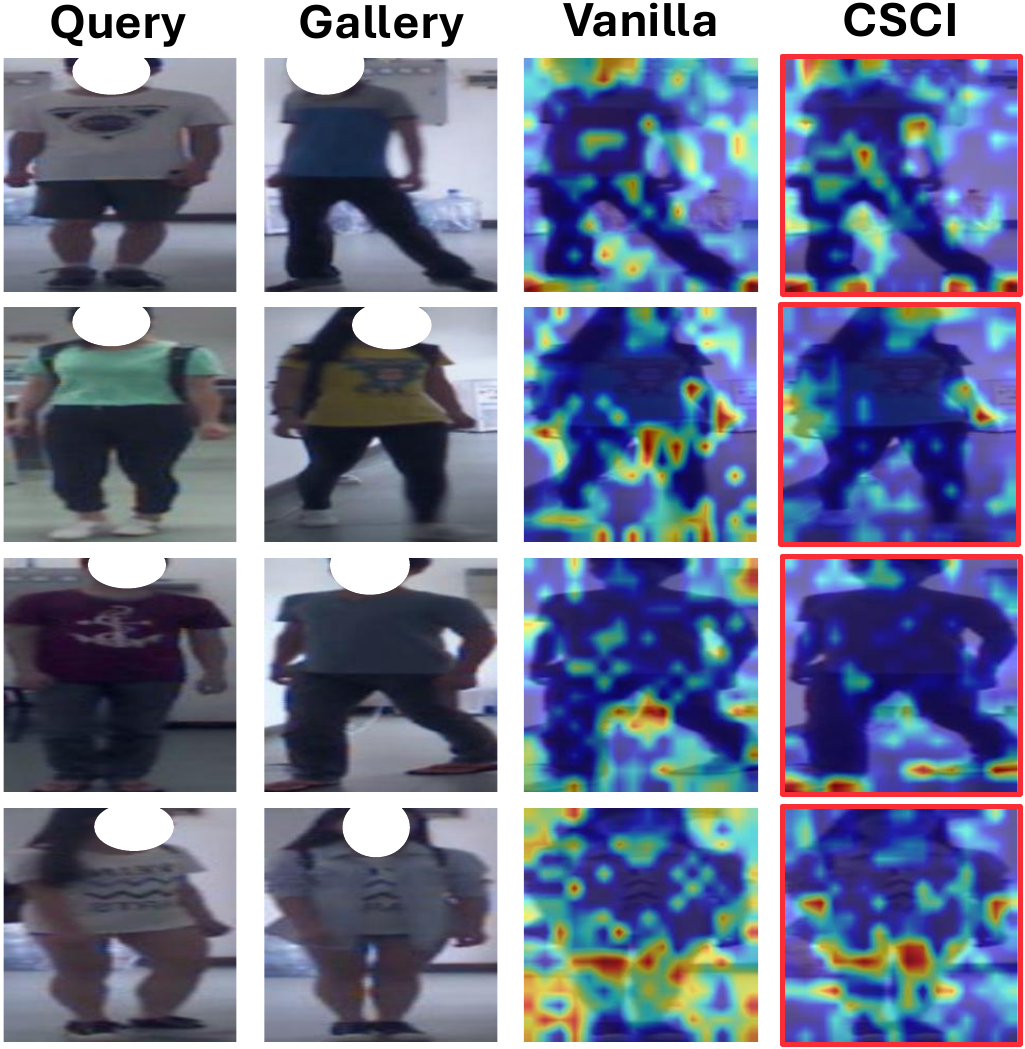} 
\caption{ \textbf{CSCI GradCAM (Different Clothing)}: Vanilla model is the baseline EVA-02 image 
 model without color tokens and CSCI (highlighted in red) is ours with EVA-02.
PRCC Images taken from a Train set, with query/gallery images similarly treated as labels for GradCAM computation. 
Under different clothing, \textbf{the model focuses on biometric features} like face, footwear, hands \etc. Vanilla baseline attention is \textbf{scatter all over the image}, compared to ours, which is more narrow and focused on the person. 
}
\label{fig:grad_cam5}
\end{figure}

\section{Intutive Explanation of S2A self-attention}
\label{sec:s2a_intutive}
S2A is the same as masked self-attention, with one key difference: ID and Color token do not influence each other’s weights. This subtle yet important difference yields 1\% gain over masked self-attention (Tab.4(a)), noteworthy as Person ReID SOTAs often differ by 1\%. 

Like masked, S2A leaks some
information between Color and ID tokens via shared spatial tokens. 
Existing methods either completely
overlap (100\% leak) color/clothes and ID \eg CAL~\cite{gu2022clothes}, CCFA\cite{han2023clothing}, 
3DInvar.~\cite{Liu_2023_ICCV}, CVSL~\cite{Nguyen_2024_WACV} or complete separation (0\% leak) via dual branches, one for colors/clothes
and the other for ID \eg  AIM~\cite{Yang_2023_CVPR}, CLIP3D~\cite{Liu_2024_CVPR} and  IRM~\cite{He_2024_CVPR}.
Dual-branch (2 transformers) is computationally impractical, while S2A achieves 2.7\% improvement over ``Traditional" complete-overlap baseline (Tab.4(a) PRCC CC R-1).
\textbf{S2A with some leak offers a middle ground btw complete and no overlap.}

\begin{figure}[!ht]
\centering
\includegraphics[width=0.98\columnwidth]{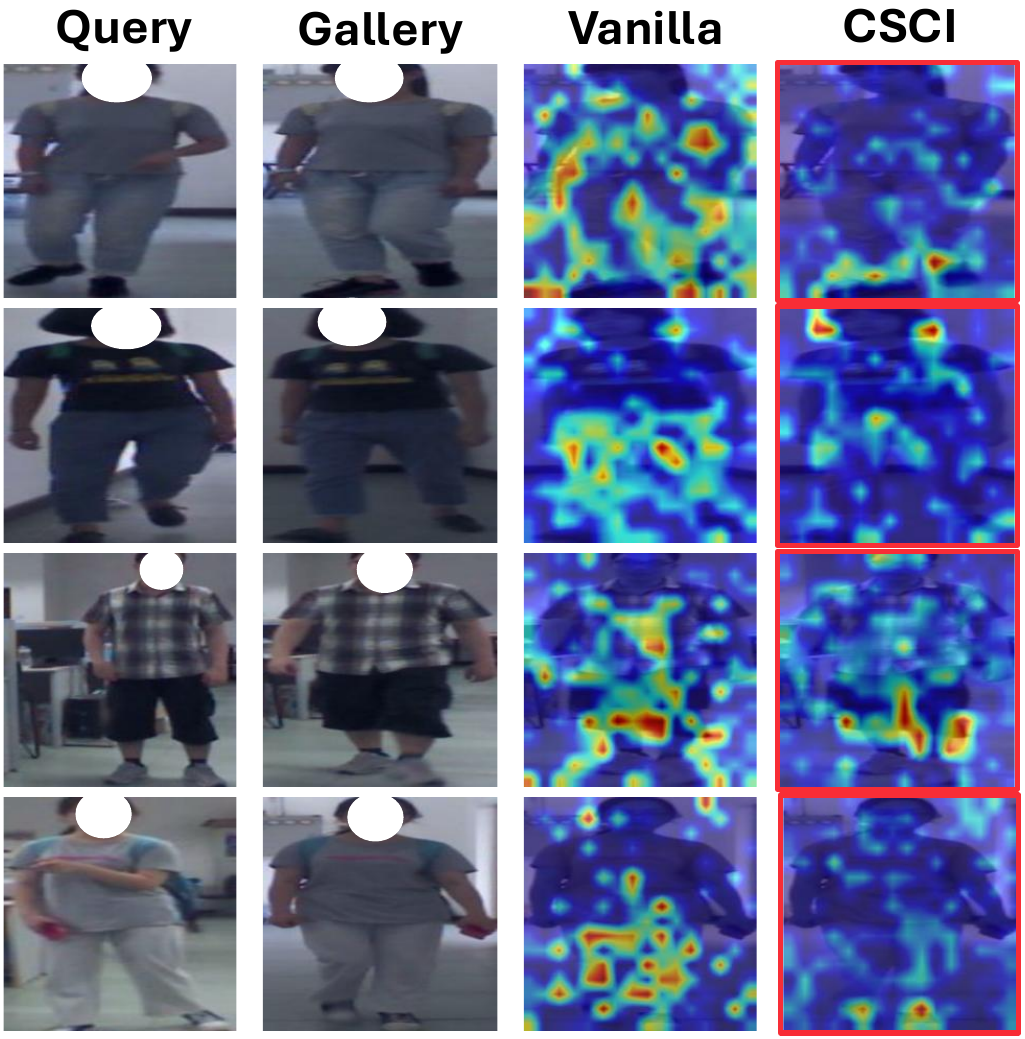} 
\caption{ \textbf{CSCI GradCAM (Similar Clothing)}: Same description as ~\cref{fig:grad_cam5}. 
Compared to different clothing, when query wears similar clothing, models tend to \textbf{focus more on the body and similar clothing}. However, CSCI in these cases selectively focuses on "hairs", and "footwear, indicating robustness against similar clothing.  
}
\label{fig:grad_cam6}
\end{figure}

\section{CSCI GradCAM}
\label{sec:grad_cam}
Grad-CAM is applied on the PRCC train set image (denoted as `Query' and `Gallery') via similarity from two images (query), with and without the similar clothing. \\
\noindent \textbf{(a) Different clothing:} \textbf{\Cref{fig:grad_cam5}} shows the attention of baseline (vanilla model) and our CSCI, where both models tend to focus on biometric features like footwear (footwear doesn't change as much as clothing), body edges, hair, and facial features. However, CSCI pays significantly less attention to backgrounds compared to the vanilla model, whose attention is scattered all over the image.  \\
\noindent \textbf{(b) Similar CLothing:} 
\textbf{\Cref{fig:grad_cam6}}
Both models here focus more on body / similar clothing, \eg pants, shirt, \etc  
However, CSCI maintains its focus on biometric areas like footwear, hair, and hands.
This helps validate the advantages of color disentanglement under similar clothing, and robustness against similar clothing.

\begin{table*}[!t]
\centering
\begin{tabular}{@{ }l@{ }|
@{ }c@{ }|
@{}P{0.82cm}@{}|@{ }P{1.8cm}@{ }|
P{0.95cm}P{0.95cm}|
P{0.95cm}P{0.95cm}||
P{0.95cm}P{0.95cm}|
@{}P{0.9cm}@{}P{0.9cm}@{ }}
\toprule
\multirow{3}{*}{Method}
 & \multirow{3}{*}{Venue} & 
\multicolumn{2}{c|}{\multirow{2}{*}{\begin{tabular}[c]{@{}c@{}} Additional \\ Attributes\end{tabular}}}
& \multicolumn{4}{c||}{LTCC} & \multicolumn{4}{c}{PRCC} \\
\cline{5-12} 
& & \multicolumn{2}{c|}{ } 
& \multicolumn{2}{c|}{CC}  & \multicolumn{2}{c||}{General} & \multicolumn{2}{c|}{CC} & \multicolumn{2}{c}{SC} \\ 
\cline{3-12}
 & & Int. & External & R-1$\uparrow$ & mAP$\uparrow$ & R-1$\uparrow$ & mAP$\uparrow$ & R-1$\uparrow$ & mAP$\uparrow$ & R-1$\uparrow$ & mAP$\uparrow$ \\

\toprule
   BSGA~\cite{bmvc_lcccpr} & BMVC'22 & CL & BP &  - & - & - & - & 61.8 & 58.7 & 99.6 & 97.3 \\ 
 CAL \cite{gu2022clothes} & CVPR'22 & CL & & 
   40.1 & 18.0 & 74.2 & 40.8 & 55.2 & 55.8 & \B{100} & \SB{99.8}\\ 
    3DInvar.~\cite{Liu_2023_ICCV}& ICCV'23 
    & \comm{CL+Co}CL & Po+BS & 40.9 & 18.9 
    & - & - & 56.5 & 57.2 & - & - \\ 
    AIM \cite{Yang_2023_CVPR} & CVPR'23 & \comm{CL+Co}CL & & 40.6 & 19.1 & 76.3 & 41.1 & 57.9 & 58.3 & \B{100} & \B{99.9} \\
    DCR.
    \cite{10036012}& TCSVT'23
    & CL & BP+Co. & 
    41.1 & 20.4 & 76.1 & 42.3 & 57.2 &  57.4 & \B{100} & 99.7 \\ 

    CCFA~\cite{han2023clothing} & CVPR'23 & \comm{CL+Co}CL & & 45.3 & 22.1 &  75.8 & 42.5 &  61.2 & 58.4 & 99.6 & 98.7 \\ 
    

CVSL~\cite{Nguyen_2024_WACV}& WACV'24 & CL & Po &  44.5 & 21.3 & 76.4 &  41.9 & 57.5 & 56.9 & 97.5 & 99.1 \\ 

CCPG~\cite{Nguyen_2024_CVPR} & CVPR'24 &CL & Po& 46.2 & 22.9 & 77.2 & 42.9 & 61.8 & 58.3 & \B{100} & 99.6\\ 

FIRe$^2$~\cite{wang2024exploring} & ITIFS'24 & - & - & 44.6  & 19.1 & 75.9 & 39.9 &  65.0 & 63.1 & 100  & 99.5  \\

\hline 
CLIP3D~\cite{Liu_2024_CVPR} & CVPR'24 & - &BS+Text & 42.1 &  21.7 & - & - &  60.6 &  59.3 &- & - \\

IRM~\cite{He_2024_CVPR}$\dagger$ & CVPR'24 & CL & Text+BP& \SB{46.7} & - & 66.7 & - & - & - & - & - \\ 
\hline 
\hline 
\textbf{ReFace+AIM} \cite{arkushin2024geff}  &  WACV'24
& - & Face & 45.7 & 20.3 & 76.3 & 42.3 & \textbf{82.5} & \textbf{64.7} & 99.8 & 99.1 \\ \hline 
\hline 
\multicolumn{2}{@{}l|}{Image \textit{(Baseline)}} 
 & - & - & 44.9	& 23.1 & 80.3 & \SB{45.9} & 61.6 & 59.0 & \B{100} & \B{99.9} \\
\multicolumn{2}{@{}l|}{CSCI - Pix. Bin \textit{(our)}} & - & - & 			
\SB{46.7}\textsuperscript{+1.8} & \SB{23.6}\textsuperscript{+0.5} & \SB{80.8}\textsuperscript{+0.5}  & \SB{45.9}  & \SB{66.6}\textsuperscript{+5.0} & {60.7}\textsuperscript{+1.7} & \B{100} & \B{99.9} \\ 

\multicolumn{2}{@{}l|}{CSCI - RGB-uv \textit{(our)}} & - & - & 
\B{47.8}\textsuperscript{+2.9} & \B{24.4}\textsuperscript{+1.3} & \B{82.6}\textsuperscript{+2.3} & \B{48.0}\textsuperscript{+2.1} & {66.2}\textsuperscript{+4.6} & \SB{61.3}\textsuperscript{+2.3} & \B{100} & \B{99.9} \\ 
\hline 
\bottomrule
\end{tabular}
\caption{\textbf{LTCC \& PRCC Results (\%)}. 
Format same as Table 1 of the main submission.
ReFace is the face-based technique applied on top of AIM to enhance the performance of features.
}
\label{tab:all_ltcc_prcc_results}
\end{table*}

\begin{table*}[!th]
\centering
\begin{tabular}{
@{}P{0.4cm}@{}|@{ }l@{}|
@{}c@{}|@{}P{0.6cm}@{}|@{}P{1.8cm}@{}|
@{}P{1.15cm}@{}P{1.15cm}@{}|
@{}P{1.15cm}@{}P{1.15cm}@{}||
@{}P{1.15cm}@{}P{1.15cm}@{}P{1.15cm}@{}P{1.15cm}@{}}
\toprule
& \multirow{3}{*}{Method} & \multirow{3}{*}{Venue} & 
\multicolumn{2}{c|}{\multirow{2}{*}{\begin{tabular}[c]{@{}c@{}} Additional \\ Attributes\end{tabular}}} & \multicolumn{4}{c|}{CCVID} & \multicolumn{4}{c}{MeVID} \\
\cline{6-13}
& & & \multicolumn{2}{c|}{} & \multicolumn{2}{c|}{CC} & \multicolumn{2}{c|}{General} & \multicolumn{4}{c}{Overall} \\ 
\cline{4-13}
& & & Int. & External & R-1$\uparrow$ & mAP$\uparrow$ & R-1$\uparrow$ & mAP$\uparrow$ & R-1$\uparrow$ & R-5$\uparrow$ & R-10$\uparrow$ & mAP$\uparrow$ \\
    \toprule
   \multirow{6}{*}{\rotatebox[]{90}{ResNet / CNN}} & CAL 
   \cite{gu2022clothes} & CVPR'22 & CL & & 81.7 & 79.6 & 82.6 & 81.3 & 52.5 & 66.5 & 73.7 & 27.1 \\ 
    & 3DInvarReID 
    \cite{Liu_2023_ICCV} & ICCV'23
    & CL & Po+BS+Co & 84.3 & 81.3 & 83.9 & 82.6 & - & - & - & - \\ 
    & DCR-ReID 
    \cite{10036012} & TCSVT'23
    & CL & BP+Co & 83.6 & 81.4 & 84.7 & 82.7 & - & - & - & - \\ 
    & SEMI 
    \cite{nguyen2024temporal} & WACV'24 
    & - & BS & 82.5 & 81.9 & 83.1 & 81.8 & - & - & - & -  \\ 
    & ShARc
    ~\cite{Zhu_2024_WACV}& WACV'24
    & - & Sil+BS+Po & 84.7 & 85.2 & 89.8 & 90.2 & \B{59.5} & \B{70.3} & \B{77.2} & \B{29.6} \\
    & LIFTCAP
    ~\cite{10497899} & TVT'24
    & CL & Co+BP+CL & 85.7 & 83.0 & 86.3 & 84.1 & - & - & - & -  \\ 
    \hline 
    \multirow{2}{*}{\rotatebox[]{90}{FM}}
    & CLIP3D
    ~\cite{Liu_2024_CVPR} & CVPR'24
    & & BS+Text & 82.4 & 83.2 & 84.5 & 83.9 & - & - & - & -  \\ 
    & GBO
    ~\cite{li2024rethinking} & ArXiv'24 
    & CL & BP+Po+Diff & 86.9 & 83.5 & 89.7 & 87.1 & - & - & - & -  \\ 
\hline 
    \multirow{4}{*}{\rotatebox[]{90}{EVA-02}}
& \multicolumn{2}{@{}l@{}|}{Image \textit{(baseline)}}& - & - & 
86.4 & 87.0 & 88.8 & 89.0 &
73.5 & 83.7	& 86.2 & 47.5 \\
& \multicolumn{2}{@{}l@{}|}{EZ-CLIP (EC) \textit{(baseline)}} & - & - & 
90.3 & 90.5 & 91.3 & 91.4 & 
\SB{76.6} & 84.8 & 87.0 & 55.5 \\
&\multicolumn{2}{@{}l@{}|}{EC + CSCI - Pix. Bin \textit{(our)}}
 &- & - & 
 \SB{90.5} & \SB{90.8}  & {91.2}  & {91.5}  & 
 \B{79.1}& \B{87.5} & \B{89.1} & \B{56.9}  \\  		
&\multicolumn{2}{@{}l@{}|}{EC + CSCI - RGB-uv \textit{(our)}}
& - & - & \B{90.8} & \B{91.3} & \SB{91.7} & \B{92.2} & 
 \B{79.1}& \SB{87.2} & \SB{88.6} & \SB{55.6}  \\  		
\hline \hline 
\multirow{3}{*}{\rotatebox[]{90}{Method}}
& {\textbf{CCVReID+CAL}
~\cite{wang2022benchmark}} & ArXiv'22 
& & Sil+Gait & 88.1 & 84.5 & 89.7  &  87.1 & - & - & - & - \\ 
& {\textbf{ReFace + CAL} 
\cite{arkushin2024geff}} & WACV'24
& - & Face & \SB{90.5} & - & 89.2 & - & - & - & - & - \\ 
& {\textbf{DCGN + CAL}
~\cite{10.1145/3643490.3661806}} & ACM'24
& & Face & 89.6 & 89.2  &  \B{92.3} & \SB{91.9} & 58.5 & - & - &  27.5  \\  
\bottomrule
\end{tabular}
\caption{\textbf{CCVID \& MEVID Results  (\%)}. 
All symbols from Table 1 and Table 2 from the main submission. The last rows highlighted in bold indicate the techniques-based approaches.}
\label{tab:all_mevid_ccvid_full_results}
\end{table*}

\begin{table*}[!th]
\centering
\begin{tabular}{@{}P{2.6cm}|@{}P{1.4cm}@{}|
@{}P{0.9cm}@{}P{0.9cm}@{}||
@{}P{0.9cm}@{}P{0.9cm}@{}|
@{}P{0.9cm}@{}P{0.9cm}@{}||
@{}P{0.9cm}@{}P{0.9cm}@{}|
@{}P{0.9cm}@{}P{0.9cm}@{}||
@{}P{0.9cm}@{}P{0.9cm}@{}|
@{}P{0.9cm}@{}P{0.9cm}@{}}
\toprule
\multirow{2}{*}{Method} & \multirow{2}{*}{Venue} & \multicolumn{2}{c|}{Overall}  & \multicolumn{2}{c|}{SC} & \multicolumn{2}{c||}{CC} & \multicolumn{2}{c|}{SL} & \multicolumn{2}{c||}{DL} & \multicolumn{2}{c|}{SS} & \multicolumn{2}{c}{DS} \\ 
\cline{3-16}
& & 
R-1$\uparrow$ & mAP$\uparrow$ &
R-1$\uparrow$ & mAP$\uparrow$ &
R-1$\uparrow$ & mAP$\uparrow$ &
R-1$\uparrow$ & mAP$\uparrow$ &
R-1$\uparrow$ & mAP$\uparrow$ &
R-1$\uparrow$ & mAP$\uparrow$ &
R-1$\uparrow$ & mAP$\uparrow$ \\
   \hline 
CAL$^*$~\cite{gu2022clothes} & CVPR & 52.5 & 27.1 & 56.6 & 39.0 & 3.5 & 4.3 & 42.1 & 24.7 & 35.0 & 22.2 & 42.3 & 24.3 & 35.2 & 20.6 \\ 
AGRL$^*$~\cite{wu2020adaptive} & IP & 48.4 & 19.1 &  51.4 & 32.6 & 4.9 & 5.7 & 27.6 & 18.1 & 41.1 & 22.5 & 40.3 & 22.1 & 29.5 & 17.7  \\
PSTA$^*$~\cite{9710753} & ICCV & 46.2 & 21.2 &  49.0 & 29.7 & 5.6 & 5.1 & 36.8 & 20.0 & 28.6 & 16.5 & 34.3 & 18.6 & 29.9 & 16.8 \\ 
Attn-CL$^*$~\cite{pathak2020video} & AAAI & 46.5 & 25.9 & 50.7 & 34.1 & 2.1 & 4.2 & 41.5 & 23.9 & 33.9 & 19.6 & 35.7 & 23.1 & 33.6 & 21.2 \\ 
ShARc~\cite{Zhu_2024_WACV}& WACV & 59.5 & 29.6 & - & -  & - & -  & - & -  & - & -  & - & -  & - & - \\ 
DCGN~\cite{10.1145/3643490}& ACM & 58.5 & 27.5 & - & - & 13.4 &  6.8 & - & - &  - & - &  - & - &  - & -    \\ 
\hline 

\multicolumn{2}{@{ }c|}{Image \textit{(Baseline)}}
 & 73.5 & 47.5 & 78.0 & 64.1 & 17.6 & 16.2 & 63.9 & 44.6 & 56.1 & 40.6 & 66.2 & 44.8 & 51.1 & \SB{47.8}\\ 

\multicolumn{2}{@{ }c|}{ EZ-CLIP (EC) \textit{(Baseline)} }
  & \SB{76.6} & 55.5 & 80.9 & \B{72.7} & 20.8 & 19.8 & 68.4 & \SB{52.7} & 59.7 & 47.0 & \SB{72.3} & \B{53.4} & 61.1 & 46.8 \\

\multicolumn{2}{@{ }c|}{ EC + CSCI - Pix. Bin \textit{(Our)} } & 
\B{79.1} & \B{56.9} & 
\B{83.7} & \B{72.7} & 
\SB{22.0} & \B{22.0} & 
\B{71.4} & \B{52.9} & 
\B{62.4} & \B{48.9} & 
\B{73.5} & \SB{53.2} & 
\B{65.9} & \B{48.4}\\ 

\multicolumn{2}{@{ }c|}{ EC + CSCI - RGB-uv \textit{(Our)} } & 
\B{79.1} & \SB{55.6} & 
\SB{83.5} & \SB{71.5} & 
\SB{22.2} & \SB{21.6} & 
\SB{69.9} & 51.6 & 
\SB{61.0} & \SB{47.5} & 
71.8 & 52.2 &
\SB{64.3} & 46.8\\ 
    \bottomrule
\end{tabular}
\caption{\textbf{MeVID Results  (\%)}. $^*$ denotes results reported with benchmarking. EC means EZ-CLIP}
\label{tab:all_metrics_mevid}
\end{table*}

\section{Loss formulations}
\label{sec:loss_formulations}

Cross entropy loss for a batch of `N' images, and classifier layer $W_{k}$ : 
\begin{equation}
    \mathcal{L}_{CE}^{ID}= -\sum_{i=1}^N log \frac{e^{W_{y_i} x_i + b_{y_i}}}{ 
    \sum_{k=1}^C e^{W_{k}x_i + b_k}
    }
\end{equation}
where $y_i$ is the label vector for $x_i$ and C classes.

 Triplet loss ($\mathcal{L}_{Tripelt}$) for a minibatch terms the i-th as "anchor" (i,A) $f_{ReID}^{iA}$.
 Samples in minibatches which label vector (identity) is same as that of anchor (i,A) are termed as `positives' (i,P) $f_{ReID}^{i,P}$. Simialrly, samples which belong to different identity that to of anchor are termed as `negatives' (i,N) $f_{ReID}^{i,N}$.

\begin{align}
    & \mathcal{L}_{Tripelt}(f_{ReID}^{iA}, f_{ReID}^{i,P} , f_{ReID}^{i,N} ) \nonumber 
    = max\bigg(0, Margin + \\
     & Dist\Big(f_{ReID}^{iA}, f_{ReID}^{i,P}\Big) - Dist\Big(f_{ReID}^{iA}, f_{ReID}^{i,N} \Big)\bigg)
\end{align}

 The regression MSE loss for predicting colors : 
 \begin{equation}
     \mathcal{L}_{MSE}^{Color} = \| f_{CO} - f_{ReID}\|^2_2
 \end{equation}

 The disentanglement loss ($\mathcal{L}_{DE}$) maximizes angular distance 
 \begin{equation}
     \mathcal{L}_{DE} = \bigg| \frac{f_{ReID}}{\|f_{ReID}\|} . \frac{f_{CO}}{\|f_{CO}\|} \bigg|
 \end{equation}

\section{All Results
\textit{(Under Comparison with SOTA Methods
in Experiments)}
}
\label{sec:all_results}

In the main paper, technique-based approaches which take existing models and improve their accuracy, mostly via facial features were skipped. We have added them here for completion. 
We have skipped the results for model agnostic technique-based methods like ReFace~\cite{arkushin2024geff} and DCGN~\cite{10.1145/3643490.3661806}.
Note these methods can be applied to our features as well. 
For completeness purposes, \cref{tab:all_ltcc_prcc_results} and \cref{tab:all_mevid_ccvid_full_results}, show results with approaches.

\section{MEVID All Protocols 
(\textit{Under Experiment Settings}) }
\label{sec:mevid_protocols}
While existing works have reported accuracy mostly on the Overall category and Clothes changing category (CC), Original work~\cite{Davila2023mevid}, evaluated baseline models on various metrics as well. 
These metrics include: 
(i) \textit{SC}: Query-Gallery wears the \underline{S}ame \underline{C}lothes
(ii) \textit{SL}: Query-Gallery recorded in the were recorded in the \underline{S}ame  \underline{L}ocation (same background).
(iii) \textit{DL}: Query-Gallery recorded in the were recorded in the \underline{D}ifferent  \underline{L}ocation.
(iv) \textit{SS}: Query-Gallery recorded in the were recorded in the \underline{S}ame Resolution/\underline{S}cale.
(v) \textit{DS}: Query-Gallery recorded in the were recorded in the \underline{D}ifferent  Resolution/\underline{S}cale.
These results are reported in the \cref{tab:all_metrics_mevid}.

\section{Limitations and Ethical Statement}
\label{sec:limitations}
Figure 5 hints at a cross-information leak between $f_{ReID}$ and $f_{CO}$.
To address this, color embeddings might need to be entirely separated from the ReID model, warranting a deeper analysis of computation overhead. 
Furthermore, the color aspect of clothing serves as a proxy for appearance bias when clothing labels are absent. 
Fine-grained clothing attributes and texture may outperform colors. 

Our ReID work relates to surveillance, which poses a risk to privacy if deployed without consent. 
Additionally, our Color token would eventually learn some color bias (Color token influences spatial tokens which in turn influences ReID features), and may start treating races/skin color differently which may be an important consideration when applying our model in the real world.  

Fig.5(a) shows that similar background/illumination causes the same clothing to be assigned to different k-means color clusters. This can be considered as a limitation as noisy color clusters or contextual benefit where appearance bias dictates the current appearance, helping remove the current appearance bias.
Additionally, Fig.7 (right) highlight EVA mismatches due to similar illumination, which indicate that a more fine-grained color may be needed (horizontal split of colors?).

\end{document}